\documentclass[12pt, letterpaper]{article}

\usepackage[utf8]{inputenc}
\usepackage[T1]{fontenc}
\usepackage{mathptmx} 
\usepackage{textcomp}
\usepackage{booktabs}
\usepackage[dvipsnames]{xcolor}

\usepackage[margin=1in]{geometry}

\usepackage{amsmath}%
\numberwithin{equation}{section}
\usepackage{amsthm}
\usepackage{amsxtra}%
\usepackage{amsfonts}%
\usepackage{amssymb}%

\usepackage{hyperref}
\usepackage{graphicx}
\usepackage{subfig}
\hypersetup{colorlinks,breaklinks,
             linkcolor=blue,urlcolor=blue,
             anchorcolor=blue,citecolor=blue}
\usepackage{xcolor}

\usepackage[T1]{fontenc}
\usepackage{booktabs}
\usepackage[margin=1in]{geometry}
\usepackage{adjustbox,lipsum}
\usepackage{threeparttable}

\usepackage{dsfont}

\usepackage[natbibapa]{apacite} 

\graphicspath{ {images/} }

\usepackage{comment}

\usepackage{algorithm}
\usepackage{algorithmic}
\usepackage{dsfont}
\usepackage[bb=boondox]{mathalfa}

\providecommand{\keywords}[1]
{
  \small
  \textbf{\textit{Keywords---}} #1
}

\usepackage{array}
\usepackage{rotating}

\begin{document}


\title{Using machine learning
on new feature sets extracted from 3D models of broken animal bones to classify fragments according to break agent \thanks{We would like to thank the National Science Foundation NSF Grant DMS-1816917 and the University of Minnesota's Department of Anthropology for funding this research. JC was partially supported by an Alfred P.~Sloan Research Fellowship and a McKnight Presidential Fellowship. Source code to reproduce all experimental results is available here: \url{https://github.com/jwcalder/MachineLearningAMAAZE}.}}
\author{Katrina Yezzi-Woodley\thanks{Department of Anthropology, University of Minnesota, \url{yezz0003@umn.edu} (corresponding author)} \and Alexander Terwilliger\thanks{School of Mathematics, University of Minnesota} \and Jiafeng Li \footnotemark[3] \and Eric Chen \thanks{Wayzata High School, Plymouth, Minnesota} \and Martha Tappen\thanks{Department of Anthropology, University of Minnesota} \and  Jeff Calder \footnotemark[3] \and Peter J.~Olver\footnotemark[3] }

\date{}

\maketitle

\begin{abstract}
\begin{small}
Distinguishing agents of bone modification at paleoanthropological sites is at the root of much of the research directed at understanding early hominin exploitation of large animal resources and the effects those subsistence behaviors had on early hominin evolution. However, current methods, particularly in the area of fracture pattern analysis as a signal of marrow exploitation, have failed to overcome equifinality. Furthermore, researchers debate the replicability and validity of current and emerging methods for analyzing bone modifications. Here we present a new approach to fracture pattern analysis aimed at distinguishing bone fragments resulting from hominin bone breakage and those produced by carnivores. This new method uses 3D models of fragmentary bone to extract a much richer dataset that is more transparent and replicable than feature sets previously used in fracture pattern analysis. Supervised machine learning algorithms are properly used to classify bone fragments according to agent of breakage with average mean accuracy of $77\%$ across tests.       
\end{small}
\end{abstract}


\section{Introduction}

Analyses of bone surface modifications and fracture patterns form the basis of substantial research focusing on early hominin subsistence patterns pertaining to the use of large animal food resources (i.e., meat and marrow). These methods are used to identify the agents of bone modification for the purpose of ascertaining the primary accumulating agent of zooarchaeological assemblages, exploring hominin-carnivore interactions, and determining hominin access order (i.e., hunting versus various forms of scavenging) \citep[e.g.][]{bartholomew1953ecology, binford1981bones, binford1984faunal, binford1985human, binford1985taphonomy, binford1986zhoukoudian, binford1988fact, blumenschine1986carcass, blumenschine1988experimental, blumenschine1989landscape, blumenschine1991marks, blumenschine1992scavenging, blumenschine1995percussion, blumenschine1994competition, blumenschine1987characteristics, pante2012validation, potts1983foraging, selvaggio1994carnivore, selvaggio1994evidence, selvaggio1998evidence, shipman1983early, shipman1986scavenging, bunn1986systematic, dominguez1997meat, dominguez2002hunting, dominguez2007five, dominguez2005cutmarked, pobiner2015new, marean1992captive, pante2012validation, plummer2016oldowan, bunn1993hunting}. These long-standing debates are founded on the premise that the use of large animal food resources was a highly influential factor in our evolution. Conversely, \citet{barr2022no} downplayed the role of meat consumption altogether, stating that there is no evidence for increased carnivory after the appearance of \textit{Homo erectus}. Clearly, debates continue on the extent to which large animal food resources informed our evolutionary past. 

We have been unable to resolve these debates because bone surface modifications and fracture patterns are subject to equifinality which has not been overcome due to the limitations of current methods. This is only exacerbated by concerns over inter- and intra-analyst error and intense disagreement among research groups about the validity of currently used methods \citep[e.g.][]{dominguez2017use, dominguez2019spilled, merritt2019don, james2015bad, harris2017trajectory}. Improving methods and ensuring that they are replicable could resolve long-standing debates over early hominin subsistence patterns at important paleoanthropological sites such as Dikika \citep{dominguez2012experimental, mcpherron2010evidence, dominguez2011reply, thompson2015taphonomy, dominguez2010configurational} and FLK Zinj \citep[see][and citations contained therein]{dominguez2014critical, pante2012validation, pante2015revalidation, parkinson2018revisiting}. 

Recently, \citet{thompson2019origins} hypothesized that scavenging for in-bone nutrients may have led to the origin of the Human Predatory Pattern, whereby humans hunt animals larger than themselves. If this is the case, then marrow exploitation may factor into major changes that happened during the Late Pliocene ($3.6-2.6$ Ma) and the Early Pleistocene ($2.6-1.8$ Ma), such as the first appearance of our genus \textit{Homo} \citep{bobe2021estimating, du2020statistical}, the first appearance of stone tools (3.3 Ma) and their subsequent technological advancements \citep{harmand20153, diez2015origin}, and geographic expansion \citep{prat2018first, zhu2018hominin}. \citeauthor{thompson2019origins} call for the development of new approaches and lines of analysis as a necessary step to successfully address these questions. Overcoming current methodological limitations has the potential to  open avenues for advancing our understanding of the evolutionary implications of large animal food resource use on our genus, \textit{Homo}. 

In response to current methodological challenges, especially as they pertain to resolving equifinality and improving replication, researchers have developed and are continuing to develop various methods for analyzing bone surface modifications and fracture patterns through approaches such as geometric morphometrics \citep[e.g.][]{yravedra2018differentiating, yravedra2017use, arriaza2017applications, courtenay2019hybrid, courtenay2019combining, courtenay20193d, mate2019application, palomeque2017pandora}, confocal profilometry \citep[e.g.][]{pante2017new, gumrukccu2018assessing, braun2016cut, schmidt2012preliminary}, and other digital data extraction methods \citep[e.g.][]{yezzi2021virtual, o2020computation, bello20113}. Many of these new methods rely on digital imaging, in particular 3D scanning, which has become another prominent avenue of research \citep[e.g.][]{yezzi2022batch, mate2019new} within the field. 

The incorporation of digital imaging and data extraction opens opportunities for using powerful computational tools such as machine learning which has been employed in bone modification studies \citep{dominguez2017revising, dominguez2018distinguishing, cifuentes2019deep, courtenay2019combining, byeon2019automated, dominguez2019successful, moclan2019classifying, pizarro2020dynamic, courtenay2019hybrid, courtenay2020obtaining, jimenez2020corrigendum, jimenez2020deep, moclan2020identifying, arriaza2021hunted, dominguez2021use}. However, most of these papers have focused on the use of machine learning for discriminating bone \emph{surface} modifications. 

The purpose of this paper is to investigate the application of machine learning methods as a means of classifying 3D meshes of bone fragments using new feature sets. Our collection of fragments consists of cervid appendicular long bones, that were broken either by hominins, using hammerstone and anvil, or by carnivores, specifically spotted hyenas. The 3D meshes were created using computed tomography (CT) and the Batch Artifact Scanning Protocol \citep{yezzi2022batch}. We introduce new feature sets that provide more detailed information about each bone fragment and thus are more useful for distinguishing agents of bone breakage, and offer more precise data extracted in a highly replicable manner using the virtual goniometer \citep{yezzi2021virtual}, Mesh Lab \citep{cignoni2008meshlab} and Python \citep{van2009python}. A small set of qualitative features were also incorporated in the data. Together, these data were input into machine learning algorithms to classify each fragment based on the agent of breakage.

We trained machine learning classification models in two different ways. First, we trained the classifiers to classify individual breaks, which we call break-level classification, and the fragment prediction is given by majority voting of the breaks for that fragment. The models are evaluated by their accuracy at classifying fragments, not individual breaks.  Second, we trained classifiers to classify the fragments directly, allowing the machine learning algorithm to determine the best way to combine information from the individual breaks. The classification accuracy for the break-level classifiers were only slightly higher than one could expect from random chance ($57.18\%-66.16\%$). On the other hand, the classification rates of the fragment-level classifiers, which incorporate the entire ensemble of breaks associated with each fragment, were substantially improved ($72.82\%- 79.27\%$) into a statistically meaningful range. 

\citet{moclan2019classifying} published the first paper to apply machine learning to distinguish agents of bone breakage using fracture pattern data \citep{moclan2019classifying}. They reported near perfect classification rates ($\ge98^{\circ}$) for some of their testing models. However, we identified several aspects of concern regarding the data they used, the way in which they used it, and their failure to follow basic machine learning protocols. First, there were inconsistencies in the manner in which the data were recorded. Furthermore, it appears that they bootstrapped their sample prior to splitting it into training and test sets; moreover, they based their analysis on global, fragment-level variables, but split the sample at the break-level, which is not permitted in a proper application of machine learning analysis because there are several breaks per fragment. Handling the data in these ways has the effect of contaminating the training set with data from the test set which, in turn, can falsely inflate success rates in classification. As we discuss in detail in Section 4, these errors effectively invalidate their published results. We revisit the \citet{moclan2019classifying} analysis using both proper machine learning protocols, and, for illustrative and explanatory purposes, their inappropriate use of bootstrapping and global variables, thereby reproducing the latter misleading and overly optimistic results. We also provide the results of an experiment on randomized data showing that both bootstrapping and break-level train-test splits can arbitrarily inflate accuracy, in many cases up to $100\%$, even when no information is present in the dataset.

When used correctly, machine learning is a powerful tool that has the potential for advancing approaches for analyzing bone modifications and subsequently improving our understanding of early human evolution. Here we demonstrate that using richer data that capture more information about the features from individual breaks -- more than has ever been captured before -- offers better discriminatory power. These methods can be easily replicated by independent research teams. By comparing our application of machine learning to that of \citet{moclan2019classifying} we exemplify the ways in which machine learning can be used effectively. Finally, we argue that classifying fragmentary bone produces higher success rates when both global, fragment level and local, break level features are used. The ability to classify individual bone fragments holds promise for improving the resolution with which paleoanthropological sites can be interpreted and furnishes more useful information for interpreting our evolutionary past.

\section{Materials and Methods}

\subsection{Our Experimental Sample}

Our experimental sample (see \hyperref[tab:inventory]{Table} \ref{tab:inventory}) consisted of 463 bone fragments (3,218 breaks) from appendicular long bones (humeri, radius-ulnae, femora, tibiae, and metapodia) that were derived from \textit{Cervus canadensis} (elk) ($n=399$ fragments) and \textit{Odocoileus virginianus} (white-tailed deer) ($n=64$ fragments). Previous researchers have concluded that metapodia are not as useful for distinguishing agents of bone breakage \citep{capaldo1994quantitative} but this is specific to fracture angles on notches as measured by a contact goniometer. Given that we used new feature sets and the more precise virtual goniometer \citep{yezzi2021virtual}, we chose to include metapodia.

\begin{center}
\begin{threeparttable}[!t]
\vspace{-3mm}
\caption{Our Experimental Sample}
\vspace{-3mm}
\label{tab:inventory}
\vskip 0.15in
\begin{small}
\begin{sc}
\begin{tabular}{lrrr}
\toprule
&{\bf HSAnv} & {\bf \textit{Crocuta crocuta}} & {\bf Total}\\ 
&{\bf Fragments (Breaks)} & {\bf Fragments (Breaks)} & {\bf Fragments (Breaks)}\\ 
\midrule
{\bf Cervus canadensis}& {\bf 275 (1651)}& {\bf 124 (987)}& {\bf 399 (2638)}\\ 
\hspace{3mm} FEM & 63 (390)& 71 (534)& 134 (924)\\ 
\hspace{3mm}  HUM & 28 (159)& 27 (241)& 55 (400)\\ 
\hspace{3mm} MTPOD& 71 (409)& 0 (0)& 71 (409)\\ 
\hspace{3mm} Unident LBSF & 0 (0)& 2 (15)& 2 (15)\\ 
\hspace{3mm} RAD-ULNA & 38 (243)& 13 (96)& 51 (339)\\ 
\hspace{3mm} TIB & 75 (450)& 11 (101)& 86 (551)\\ 
{\bf Odocoileus virginianus}& {\bf 0 (0)}& {\bf 64 (580)}& {\bf 64 (580)}\\ 
\hspace{3mm} MTPOD & 0 (0)& 64 (580)& 64 (580)\\ 
\midrule
{\bf Total}& {\bf 275 (1651)}& {\bf 188 (1567)}& {\bf 463 (3218)}\\ 
\bottomrule
\end{tabular}
\end{sc}
\end{small}
\begin{tablenotes}[para,flushleft]
\footnotesize{Abbreviations: femur (FEM), humerus (HUM), metapodial (MTPOD), unidentified long bone shaft fragment (Unindent LBSF), radius-ulna (RAD-ULN), tibia (TIB), hammerstone and anvil (HSAnv).}\\
\vskip -0.1in
\end{tablenotes}
\vskip -0.1in
\end{threeparttable}
\end{center}
\medskip

Of the 463 fragments, 275 (1,651 breaks)  were produced by hominins and 188 (1,567 breaks) were produced by carnivores. 
Hammerstone and anvil were used to break the bones for the hominin sample. The carnivore sample was created by \textit{Crocuta crocuta} (spotted hyenas) at the Milwaukee County Zoo (Wisconsin) and the Irvine Park Zoo in Chippewa Falls (Wisconsin). In some instances, articulated limbs were fed to the hyenas which resulted in two fragments that could not be identified to skeletal element. \citep[See][for details regarding the experimental protocols.]{coil2017new} 

Fragments were scanned via computed tomography (CT) using the streamlined Batch Artifact Scanning Protocol \citep{yezzi2022batch} that we developed to acquire 3D models of each of the fragments, which were stored as \emph{.ply} files. Data were extracted from the 3D models of the bone fragments manually through the Graphical User Interface in Meshlab and automatically using Python scripts.

\begin{figure}[!t]
\centering
\captionsetup{margin=2cm}
\includegraphics[width=.75\textwidth]{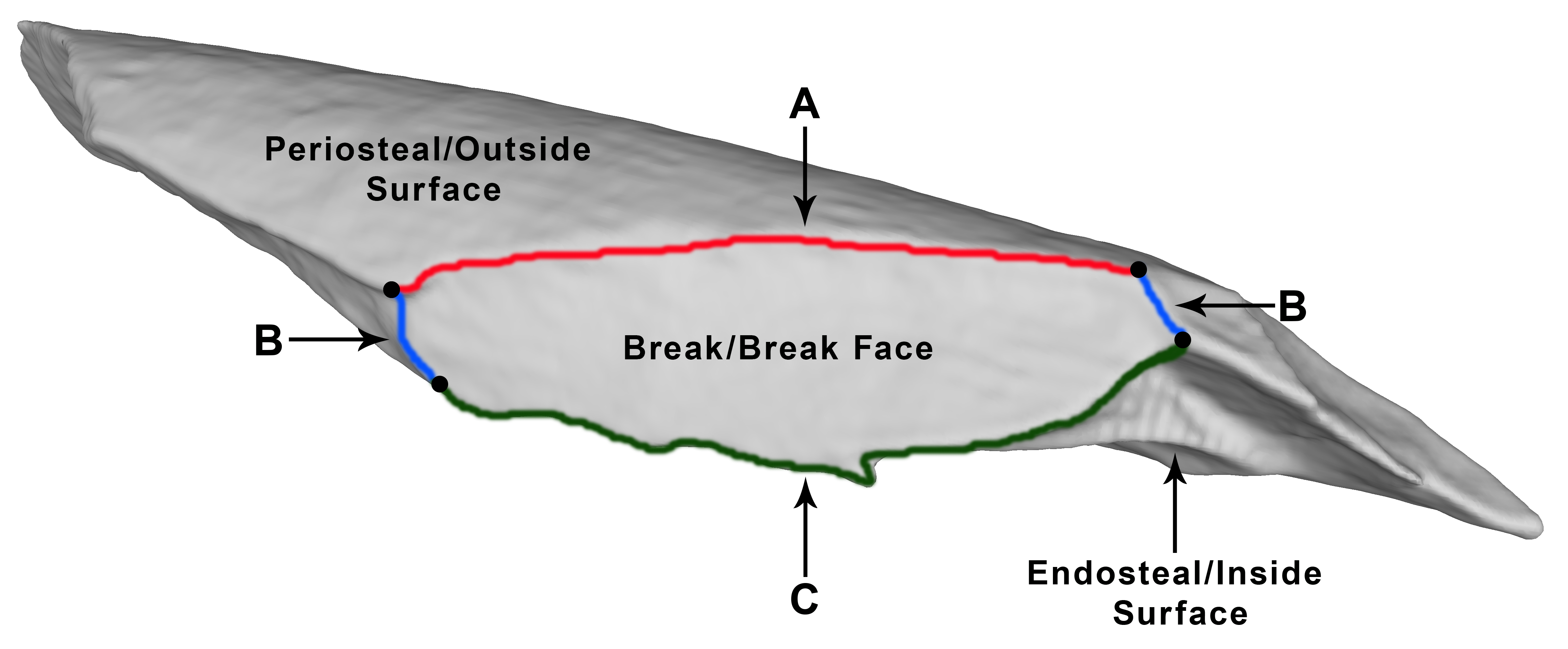}
\caption[Fragment Features]
{Fragment Features \medskip \par \small Fracture (or break) ridges are used to delineate individual breaks. One fracture ridge separates the natural outside surface of the bone from the break surface (A) Two ridges on either side of the break serve as boundaries between adjacent breaks (B). The interior fracture ridge separates the break from the natural interior surface of the bone and, in some cases, other breaks (C). }
\label{fig:anatomy}
\centering
\end{figure}

\begin{figure}[!t]
\centering
\captionsetup{margin=2cm}
\includegraphics[width=.75\textwidth]{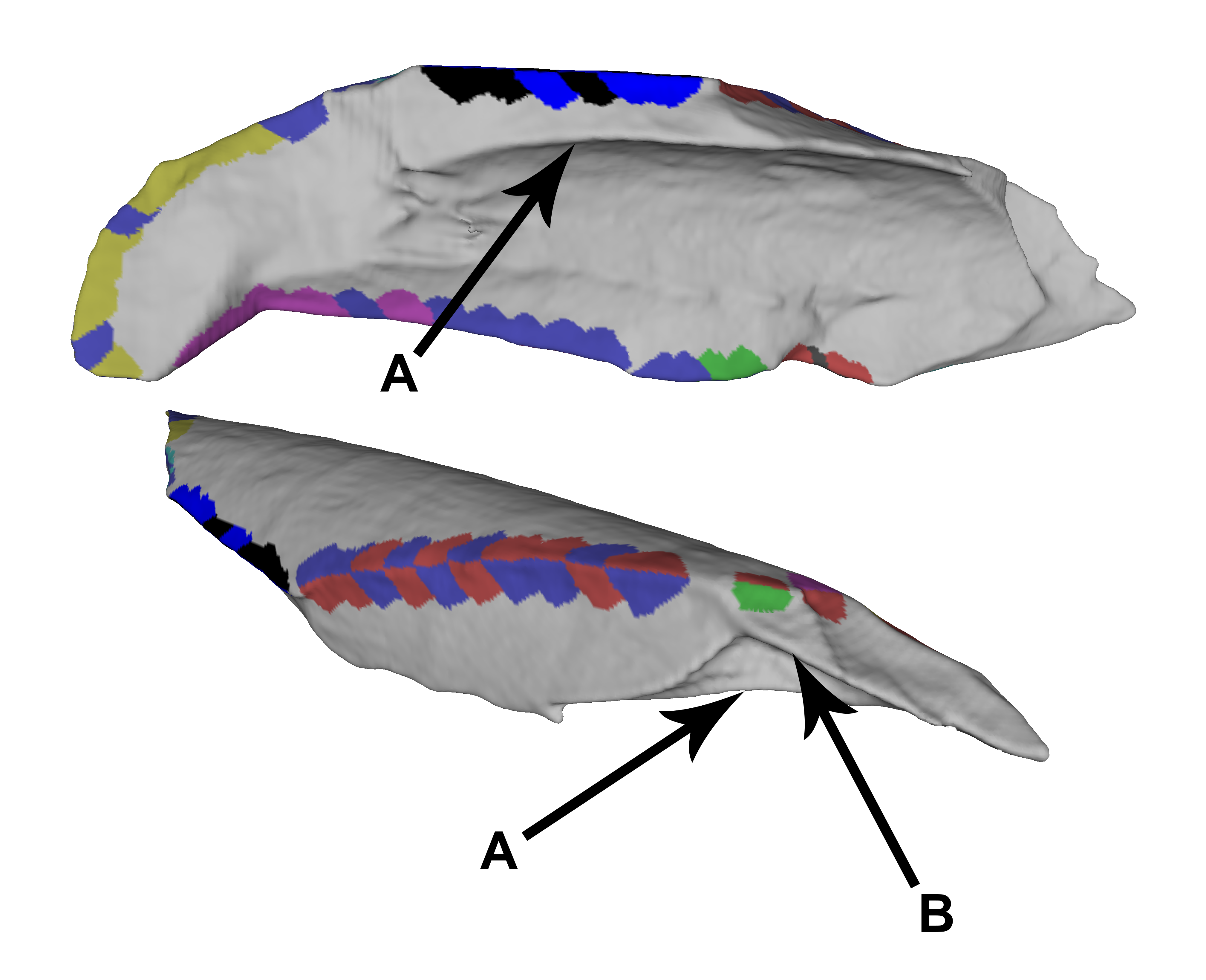}
\caption[The Interior Ridge]
{The Interior Ridge \medskip \par \small The interior edge of some breaks border the endosteal surface (A) while others border another break (B).}
\label{internal}
\centering
\end{figure}

\begin{figure}[!t]
\centering
\captionsetup{margin=2cm}
\includegraphics[width=.75\textwidth]{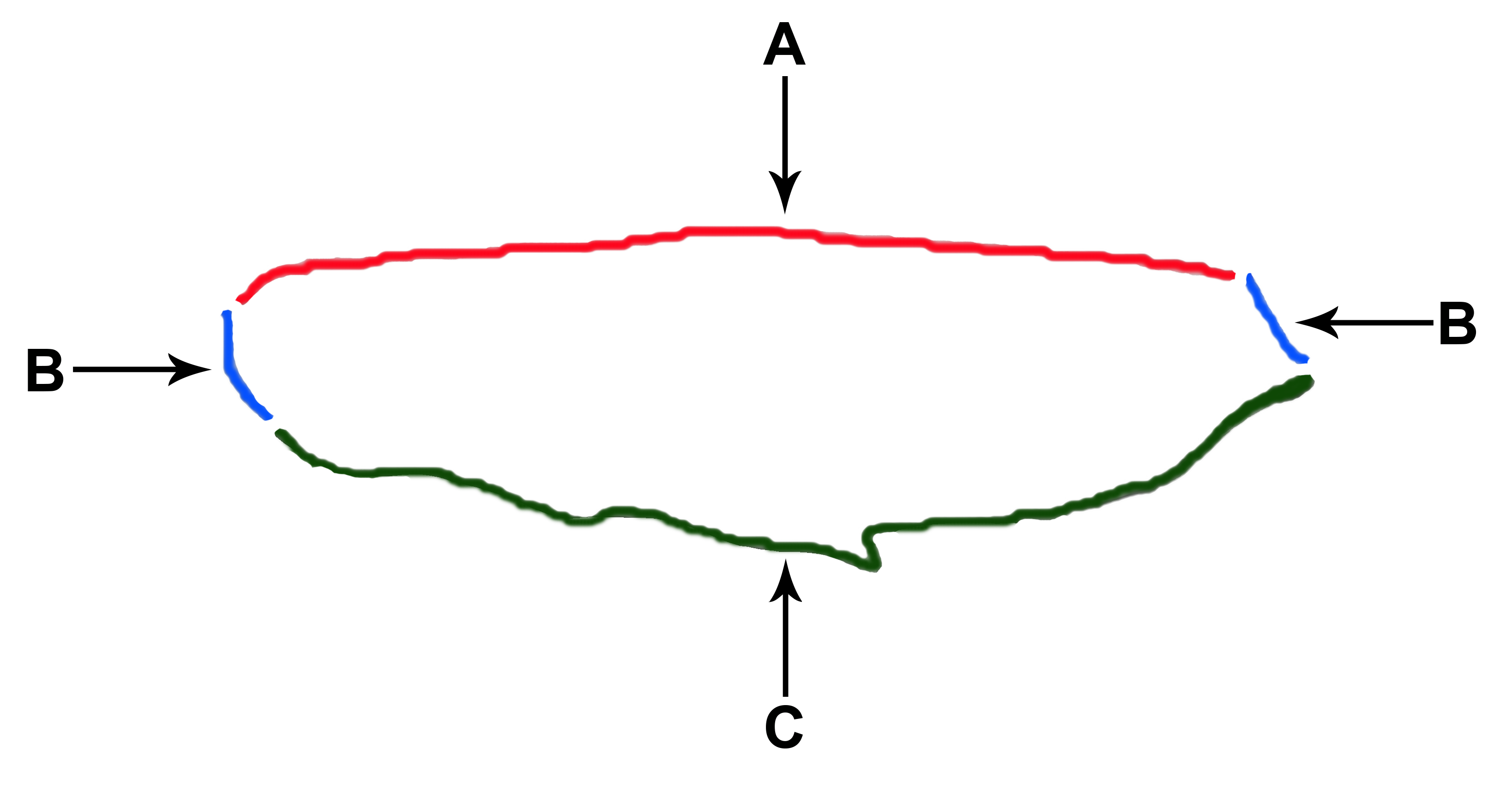}
\caption[Break Curves]
{Break Curves \medskip \par \small Break (or fracture) curves can be extracted from the fracture ridges surrounding a break face. These break curves correspond to the break ridges illustrated in \hyperref[fig:anatomy]{Figure} \ref{fig:anatomy}. The exterior curve separates the periosteal surface from the break surface (A) Two curves separate the break from adjacent breaks (B). One curve is the interior break curve (C). }
\label{fig:curves}
\centering
\end{figure}

To know how each feature was extracted from the fragments, it is important to understand how we defined the features and in particular how we differentiated breaks, because, as researchers have previously acknowledged, identifying breaks is not always a straight forward task \citep{biddick1975quantifying, davis1985taphonomic, pickering2005contribution, bunn1982meat, bunn1989diagnosing}. As \citet[p.43]{bunn1982meat} points out, the boundary between breaks is neither well-defined nor is it always obvious. Given a long bone, there is the natural outside (periosteal) surface of the bone and the natural inside (endosteal) surface of the bone. Fragments from broken long bones also have surfaces that expose what is referred to as breaks or break faces. Each break is surrounded by fracture (or break) ridges. One of those ridges constitutes the boundary between the periosteal surface and the break surface for that break (see \hyperref[fig:anatomy]{Figure} \ref{fig:anatomy}\hyperref[fig:anatomy]{A}). Two of the ridges connect adjacent break faces (see \hyperref[fig:anatomy]{Figure} \ref{fig:anatomy}\hyperref[fig:anatomy]{B}). The fourth ridge is the boundary between the break face and the endosteal surface of the bone fragment(see \hyperref[fig:anatomy]{Figure} \ref{fig:anatomy}\hyperref[fig:anatomy]{C}). In cases where the break face overlaps another break face(s) without extending through the entire thickness of the bone, the fourth ridge serves as a boundary between this break and the more internal break(s) (see \hyperref[internal]{Figure} \ref{internal}). Break curves can be extracted from those ridges (see \hyperref[fig:curves]{Figure} \ref{fig:curves}). In more complicated scenarios, the boundary of the break face may contain additional ridges of various types.  It should be noted that ridges are not always easily detectable and one must decide at which scale to accept a ridge as a boundary and break surface as a separate face. Given the precision of the tools we used to extract data, we were able to accept small breaks as separate breaks. Though further discussion is necessary in the field to standardize the definition of an individual break, the extraction methods used here provide sufficient detail in the data to begin such research.

\begin{figure}[!t]
\centering
\captionsetup{margin=2cm}
\includegraphics[width=.75\textwidth]{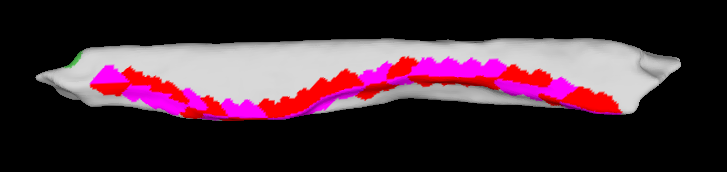}
\caption[Angle Measurements Along Ridge]
{Angle Measurements Along Ridge \medskip \par \small Angle measurements were taken along the entire fracture ridge between the natural outside surface of the bone and the break surface.}
\label{anglesedge}
\centering
\end{figure}

\begin{figure}[!t]
\centering
\captionsetup{margin=2cm}
\includegraphics[width=.75\textwidth]{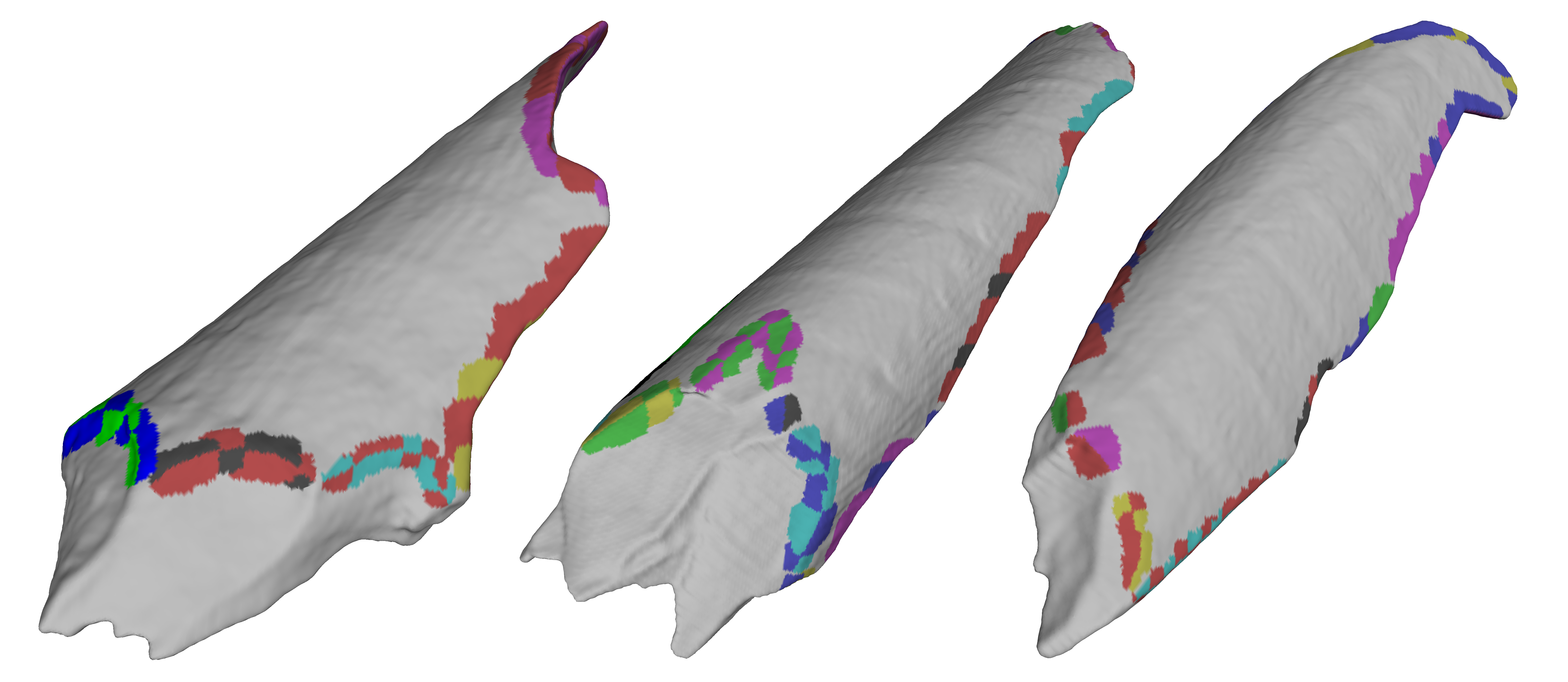}
\caption[The Virtual Goniometer Captures More Detail]
{The Virtual Goniometer Captures More Detail \medskip \par \small The virtual goniometer makes it possible to capture more information with a higher degree of detail. This includes the ability to measure small breaks. When transitioning from one break to the next, the colors of the patches where the angle measurements are taken change.}
\label{moredetail}
\centering
\end{figure}

Angle measurements were taken along the exterior fracture ridge of each break on each fragment with the virtual goniometer \citep{yezzi2021virtual} using radii between $1-3$ using geodesic distance based on the size of the break (See \hyperref[anglesedge]{Figure} \ref{anglesedge}). When an angle measurement is taken, a colorized patch appears on the mesh. The center of subsequent angle measurement was placed on the fracture ridge at the border of the colorized patch indicated by the previous angle measurement. Therefore, measurements were $1-3$ geodesic units apart. When shifting to the next break, the patch colors change (see \hyperref[moredetail]{Figure} \ref{moredetail}). The endpoints of each break were chosen using the ``Picked Points'' tool in Meshlab. The angle and endpoint data were used to calculate the arc angle (see \hyperref[arc]{Figure} \ref{arc}) and break length variables (see \hyperref[length]{Figure} \ref{length}).

Additional variables were collected manually through observation of the models such as the presence or absence of notches (see \hyperref[notch]{Figure} \ref{notch}) or trabecula. We ran the models through a Python script to extract more global features such as the mesh volume and surface area. Data were recorded and calculated  for each break, referred to here as break level data, and for the entire fragment, referred to here as fragment level data. A summary of these variables are as follows: \\

\subsubsection*{Break Level Variables:}

\begin{figure}[!t]
\centering
\captionsetup{margin=2cm}
\includegraphics[width=.75\textwidth]{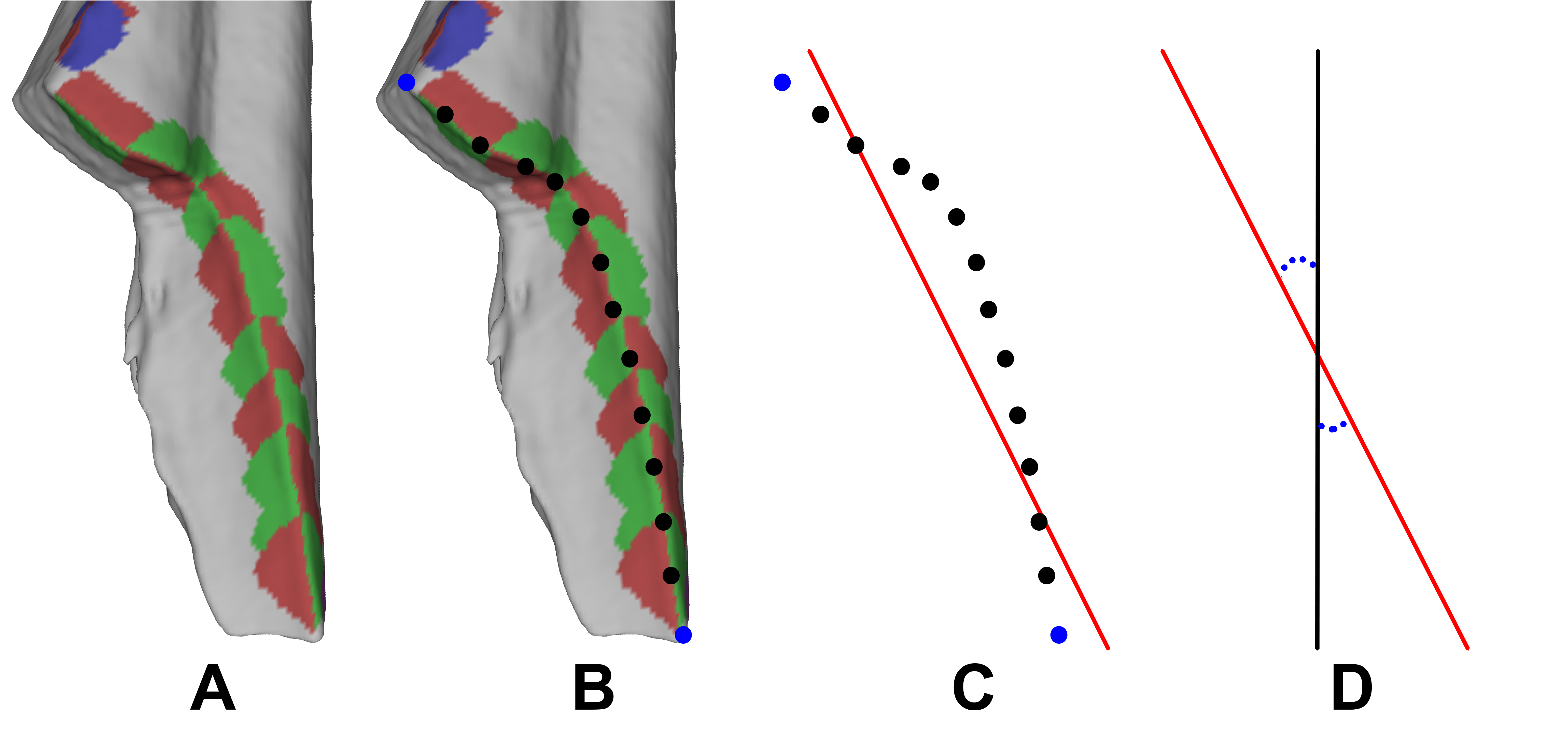}
\caption[Arc Angle]
{Arc Angle \medskip \par \small For each break curve (A) the $x-$, $y-$, and $z-coordinates$ for both the endpoints (blue circles) and the locations of each angle measurement (black points) were recorded (B). The points were used to define a best fit line (C). The angle between the best fit line and the principal axis of the fragments was recorded (D).}
\label{arc}
\centering
\end{figure}

\begin{figure}[!t]
\centering
\captionsetup{margin=2cm}
\includegraphics[width=.75\textwidth]{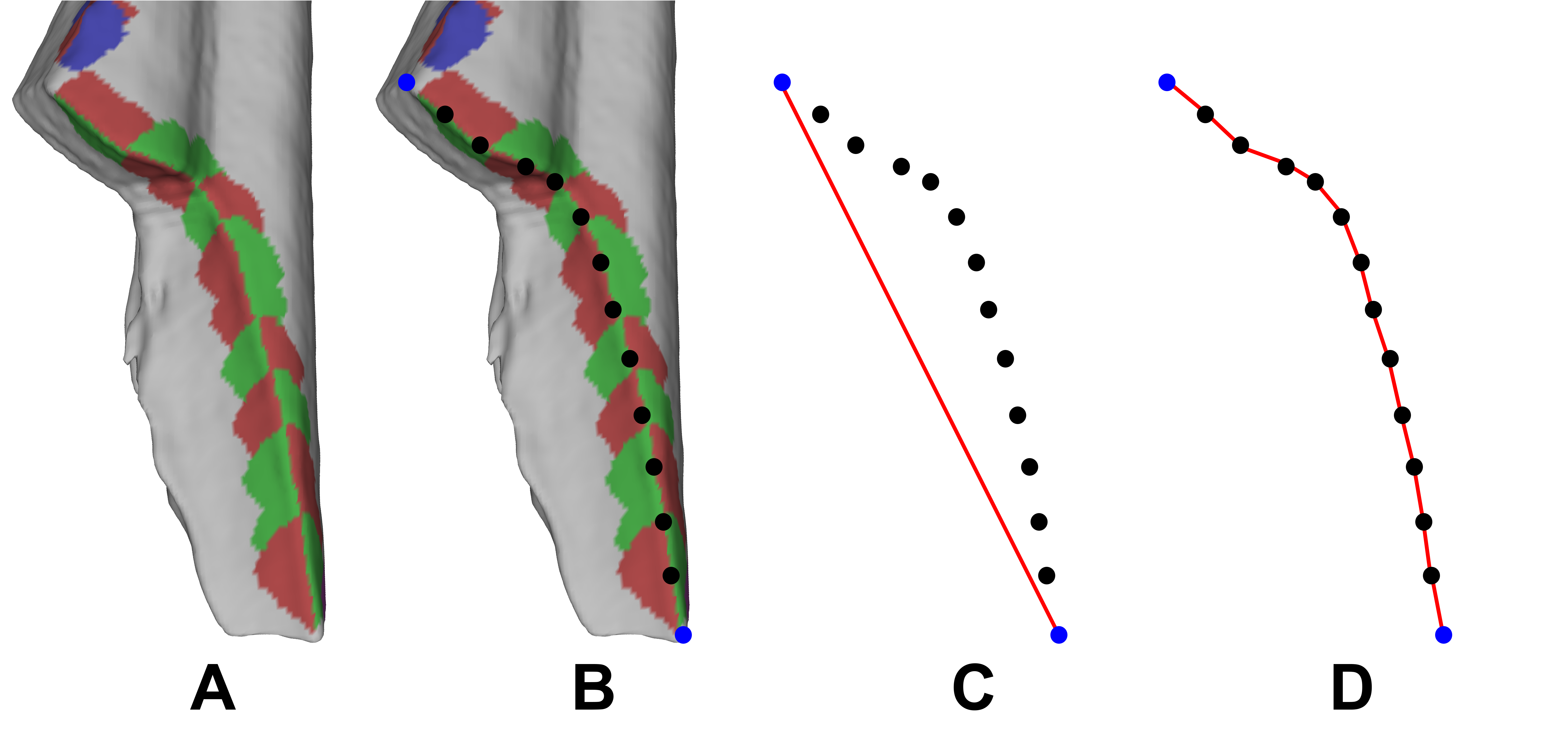}
\caption[Break Lengths]
{Break Lengths \medskip \par \small For each break curve (A) the $x-$, $y-$, and $z-coordinates$ for both the endpoints (blue circles) and the locations of each angle measurement (black points) were recorded (B). The straight line (Euclidean) distance was measured between the endpoints of the break curve (C) and the arc length was measured using all the points (D).}
\label{length}
\centering
\end{figure}

\begin{figure}[!t]
\centering
\captionsetup{margin=2cm}
\includegraphics[width=.75\textwidth]{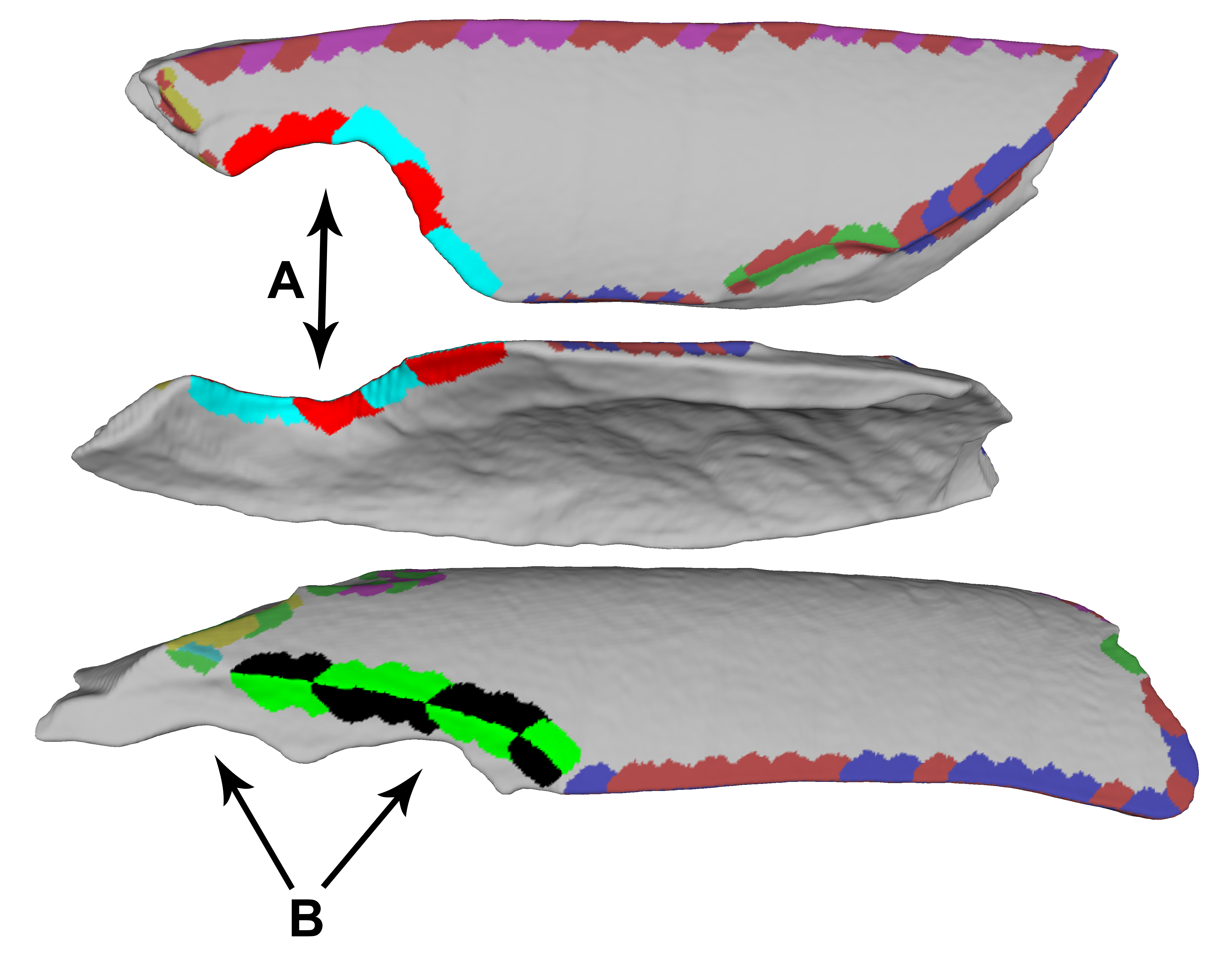}
\caption[Notches]
{Notches \medskip \par \small Notches are arcuate indentations on the bone created at the location of direct impact between the bone and the object used to break it. Some of the breaks had notches on the exterior fracture ridge (A). Some notches were found on the interior fracture ridge (B)}
\label{notch}
\centering
\end{figure}

\begin{enumerate}
    \item \textbf{Number of Angles:} This refers to the number of fracture angle measurements per break. Fracture angle measurements were taken along each break curve. A minimum of one fracture angle measurement was recorded for each break. Type: natural number
    \item \textbf{Angle data:} Because more than one angle measurement could be taken on each break curve, summary statistics were calculated for the angle measurements. This included the minimum, maximum, mean, median, standard deviation, and range. Type: continuous
    \item \textbf{Interior Edge:} If the interior ridge of the break face transitioned to another break face it was categorized as "break". If it adjoined the endosteal surface, then it was categorized as "endosteal" (see \hyperref[internal]{Figure} \ref{internal}). Type: Boolean
    \item \textbf{Interrupted:} This is a TRUE/FALSE variable indicating whether or not the break curve was interrupted by another break. Type: Boolean
    \item \textbf{Ridge Notch:} If the fracture ridge exhibited the arcuate indentation characteristic of a notch, then the break was classified as "TRUE" (see \hyperref[notch]{Figure} \ref{notch}\hyperref[notch]{A}). Type: Boolean 
    \item \textbf{Interior Notch:} If the interior ridge of the break face exhibited the arcuate indentation(s) characteristic of a notch(es), then the break was classified as "TRUE" (see \hyperref[notch]{Figure} \ref{notch}\hyperref[notch]{B}). Type: Boolean
    \item \textbf{Break Lengths:} Two measures of break length were recorded. We calculated the Euclidean distance between the two endpoints of the break curve (see \hyperref[length]{Figure} \ref{length}\hyperref[length]{C}). When using the virtual goniometer, the location of each angle measurement is automatically recorded. Using those points in conjunction with the endpoints, we calculated the arc length of each break curve (see \hyperref[length]{Figure} \ref{length}\hyperref[length]{D}). Type: continuous
    \item \textbf{Arc Angle:} This is a calculation that we used in lieu of break plane, as defined by \citet{gifford1989ethnographic}. We did this by calculating a best fit line to the ordered points along the break curve and then calculating the angle between the best fit line and the principal axis of the bone fragment. Again, the points were taken from the selected endpoints and the virtual goniometer data (see \hyperref[arc]{Figure} \ref{arc}). Type: continuous\\
\end{enumerate}

\subsubsection*{Fragment Level Variables:}

\begin{enumerate}
    \item \textbf{Number of Breaks:} We recorded the number of break faces per fragment. Type: natural number
    \item \textbf{Trabecula:} If there was trabecular bone on the fragment it was categorized as "TRUE". Type: Boolean
    \item \textbf{Volume:} The volume of the domain bounded by the surface mesh was extracted in Python. Type: continuous
    \item \textbf{Surface Area:} The surface area of the mesh was extracted in Python. Type: continuous
    \item \textbf{Bounding Box Dimensions:} The bounding box dimensions were extracted using Python. This can be thought of as the fragment length, width, and depth. Type: continuous 
    \item \textbf{Angle Data:} The summary statistics were calculated from the summary statistics of the fracture angle data calculated at the break level. We chose to do this as opposed to summarizing the original angle data because we did not want each individual angle measurement to be weighted equally. We wanted the angle data to be weighted by how the angles were summarized for each break. Type: continuous
    \item \textbf{Interior Edge:} The number of break faces with interior edges adjacent to another break were tallied per fragment as were those that were adjacent to the endosteal surface. Type: natural number
    \item \textbf{Interrupted:} The number of break faces that were interrupted were tallied per fragment. Type: natural number
    \item \textbf{Ridge Notch:} The number of break faces with fracture ridges classified as notches were tallied. Type: natural number
    \item \textbf{Interior Notch:} The number of break faces that had notching on their interior ridge were tallied. Type: natural number
    \item \textbf{Break Lengths:} Summary statistics were calculated for both measures of break curve length (Euclidean distance and arc length). Type: continuous
    \item \textbf{Arc Angle:} Summary statistics were calculated for the arc angles of the break curves. Type: continuous.
\end{enumerate}
    
\subsection{Methods}

Bone fragments were categorized using $7$ different machine learning algorithms: random forest, linear support vector machine, support vector machine using the radial basis function, neural network, linear discriminant analysis, Gaussian naive Bayes, and $k$-nearest neighbor. 
High level descriptions of the machine learning methods we used here can be found in \citet[Chapter 4]{yezzi2022thesis}.  For more detailed information about classical machine learning methods, we refer the reader to \citet{bishop2006pattern}, and for more information about deep learning and neural networks, we refer to \citet{lecun2015deep}. The code for all our experimental results can be found on GitHub.\footnote{Source Code: \url{https://github.com/jwcalder/MachineLearningAMAAZE}} 

Data were split into training (75\%) and testing (25\%) sets. This was done at the fragment level for all tests so as to avoid contaminating the training set with data from the test set. This means that for the break level tests, 25\% of the fragments were marked for the test set, and the test set was then populated by those fragment's breaks, ensuring that all breaks from a single fragment were either in the training set or the testing set. Because the train-test split was done at the fragment level, when classifying breaks, the breaks voted on which labels each fragment should receive based on the statistical mode of their predicted labels. Ties were broken at random. The accuracy reported for break-level tests is therefore the percentage of fragments that were assigned the correct label by their breaks for that algorithm in question. We  emphasize  that we \emph{never} use information from the testing set when training any of the machine learning algorithms. As is standard in machine learning, the testing set must be kept independent of \emph{all steps} in the model training procedure, so that the testing accuracy can give an unbiased evaluation of model performance on new data that was not seen during training.

Each test was repeated $300$ times with a new train and test set computed from the original data-set. The mean accuracy across all repetitions was recorded as well as the standard deviation. 

\section{Our Results}

The results of classifying bone fragments using break-level classifiers were only slightly above what can be expected from random chance ($50\%$). The mean accuracy ranged from $57.18\%-68.34\%$ with standard deviations ranging from $4.22\%-5.06\%$ (see \hyperref[tab:AMAAZEbreakresults]{Table} \ref{tab:AMAAZEbreakresults}). In particular, the mean accuracies are all lower than the $69.8\%$ accuracy we report from unsupervised learning in \hyperref[sec:un]{Section} \ref{sec:un} below, indicating that there is very little information useful for classification in the break-level dataset.

\begin{center}
\begin{threeparttable}[t!]
\vspace{-3mm}
\caption{Machine Learning Results \\ (Break Level)}
\vspace{-3mm}
\label{tab:AMAAZEbreakresults}
\vskip 0.15in
\begin{small}
\begin{sc}
\begin{tabular}{lcc}
\toprule
 & Mean  & Standard  \\
Algorithm &  Accuracy &  Deviation \\
\hline
Random forest (RF) & $68.34\%$ & $4.24\%$  \\
Support vector machine (SVM) – linear & $62.60\%$ & $4.46\%$  \\
Support vector machine - RBF & $66.16\%$ & $4.22\%$  \\
Neural network (NN) & $65.30\%$ & $5.06\%$  \\
Linear discriminant analysis (LDA)  & $64.47\%$ & $4.30\%$ \\
Gaussian naive Bayes (GNB) & $57.18\%$ & $4.76\%$  \\
$k$-nearest neighbor (KNN) & $65.19\%$ & $4.23\%$  \\
\bottomrule
\end{tabular}
\begin{tablenotes}[para,flushleft]
\item \footnotesize{300 repetitions}\\
\end{tablenotes}
\end{sc}
\end{small}
\vskip -0.1in
\end{threeparttable}
\end{center}

On the other hand, when we trained the machine learning classifiers at the fragment-level, giving the models access to summary statistics about each fragment's constituent breaks, the classification accuracy improved substantially. The mean accuracy across tests ranged from $72.82\%-79.27\%$ with lower standard deviations ($3.42\%-3.94\%$) (see \hyperref[tab:AMAAZEfragmentresults]{Table} \ref{tab:AMAAZEfragmentresults}). These results are substantially higher than the unsupervised results ($69.8\%$) in \hyperref[sec:un]{Section} \ref{sec:un}, indicating that the machine learning methods are learning from the labeled information in a significant way.

\begin{center}
\begin{threeparttable}[t!]
\vspace{-3mm}
\caption{Machine Learning Results \\ (Fragment Level)}
\vspace{-3mm}
\label{tab:AMAAZEfragmentresults}
\vskip 0.15in
\begin{small}
\begin{sc}
\begin{tabular}{lcc}
\toprule
 & Mean  & Standard  \\
Algorithm &  Accuracy &  Deviation \\
\hline
Random forest (RF) & $77.18\%$ & $3.51\%$ \\
Support vector machine (SVM) – linear & $77.24\%$ & $3.48\%$ \\
Support vector machine - RBF & $79.27\%$ & $3.42\%$ \\
Neural network (NN) & $77.95\%$ & $3.56\%$ \\
Linear discriminant analysis (LDA)  & $76.19\%$ & $3.59\%$ \\
Gaussian naive Bayes (GNB) & $72.82\%$ & $3.94\%$ \\
$k$-nearest neighbor ($k$-NN) & $77.61\%$ & $3.51\%$ \\
\bottomrule
\end{tabular}
\begin{tablenotes}[para,flushleft]
\item \footnotesize{300 repetitions}\\
\end{tablenotes}
\end{sc}
\end{small}
\vskip -0.1in
\end{threeparttable}
\end{center}

We did not tune hyperparameters for any of the methods. Hyperparameter optimization, with an appropriately chosen validation set, could have the potential to slightly improve results.  For $k$-nearest neighbor, we fixed the value $k = 25$, which worked well for our replication of \citealt{moclan2019classifying} discussed below. For the neural network, we used a fully connected neural network with three hidden layers with $100$, $1000$, and $5000$ hidden nodes in each layer, respectively. We trained the network with the Adadelta \citep{zeiler2012adadelta} optimizer with  batch size of 32, initial learning rate of 1 with scheduled decreases by $10\%$ every epoch, and we trained the network over 100 epochs. To prevent overfitting, we used dropout layers \citep{srivastava2014dropout} with dropout rate of $0.4$ between the hidden layers of the neural network. We refer the reader to \cite{lecun2015deep} for more details on deep learning.


\subsection{Unsupervised Learning}
\label{sec:un}


\begin{figure}[!t]
\centering
\captionsetup{margin=2cm}
\subfloat[Spectral Embedding]{\includegraphics[width=0.49\textwidth,clip=true,trim= 20 15 30 30]{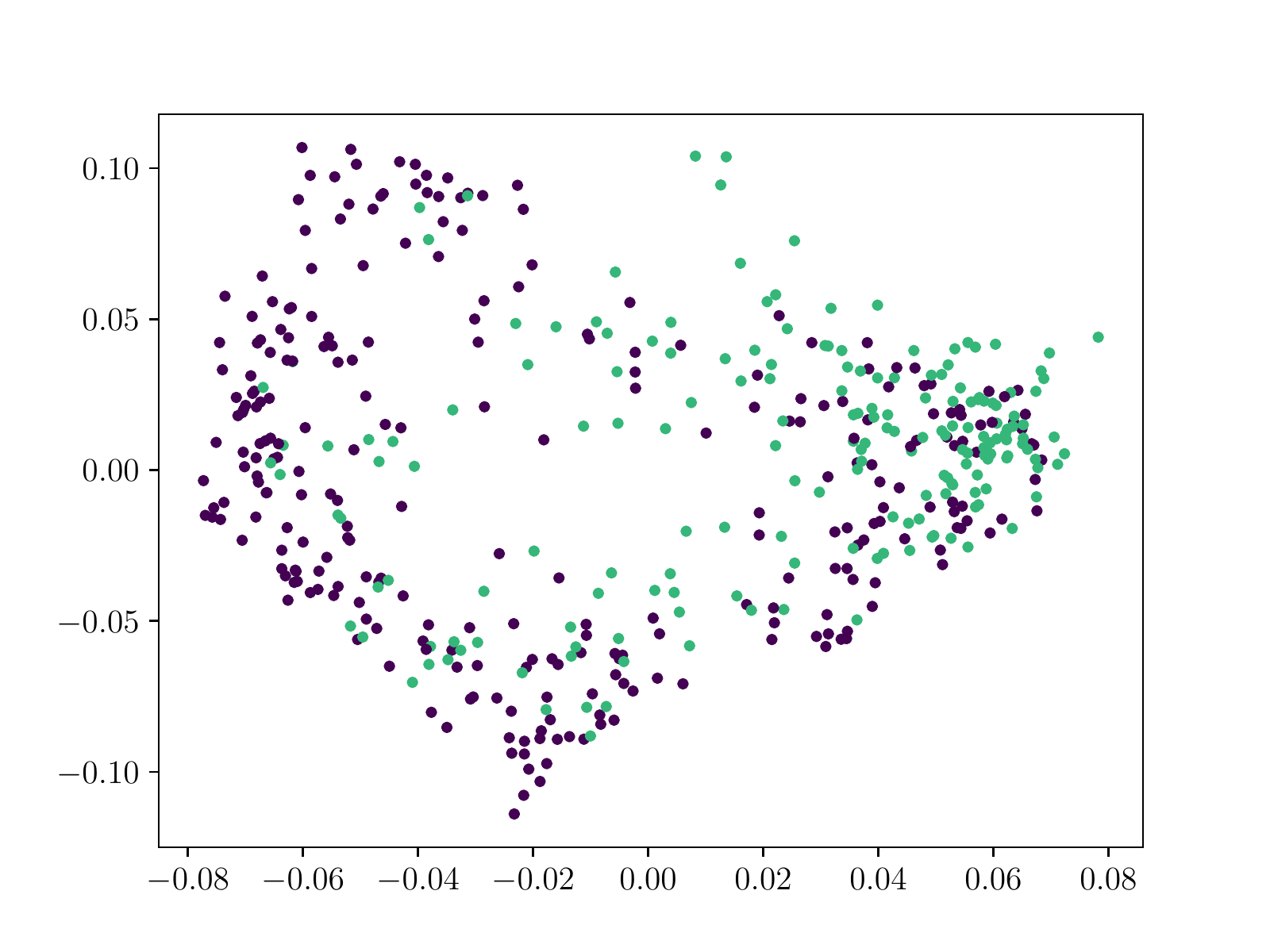} \label{fig:spec_embedding}}
\subfloat[Spectral Clustering]{\includegraphics[width=0.49\textwidth,clip=true,trim= 20 15 30 30]{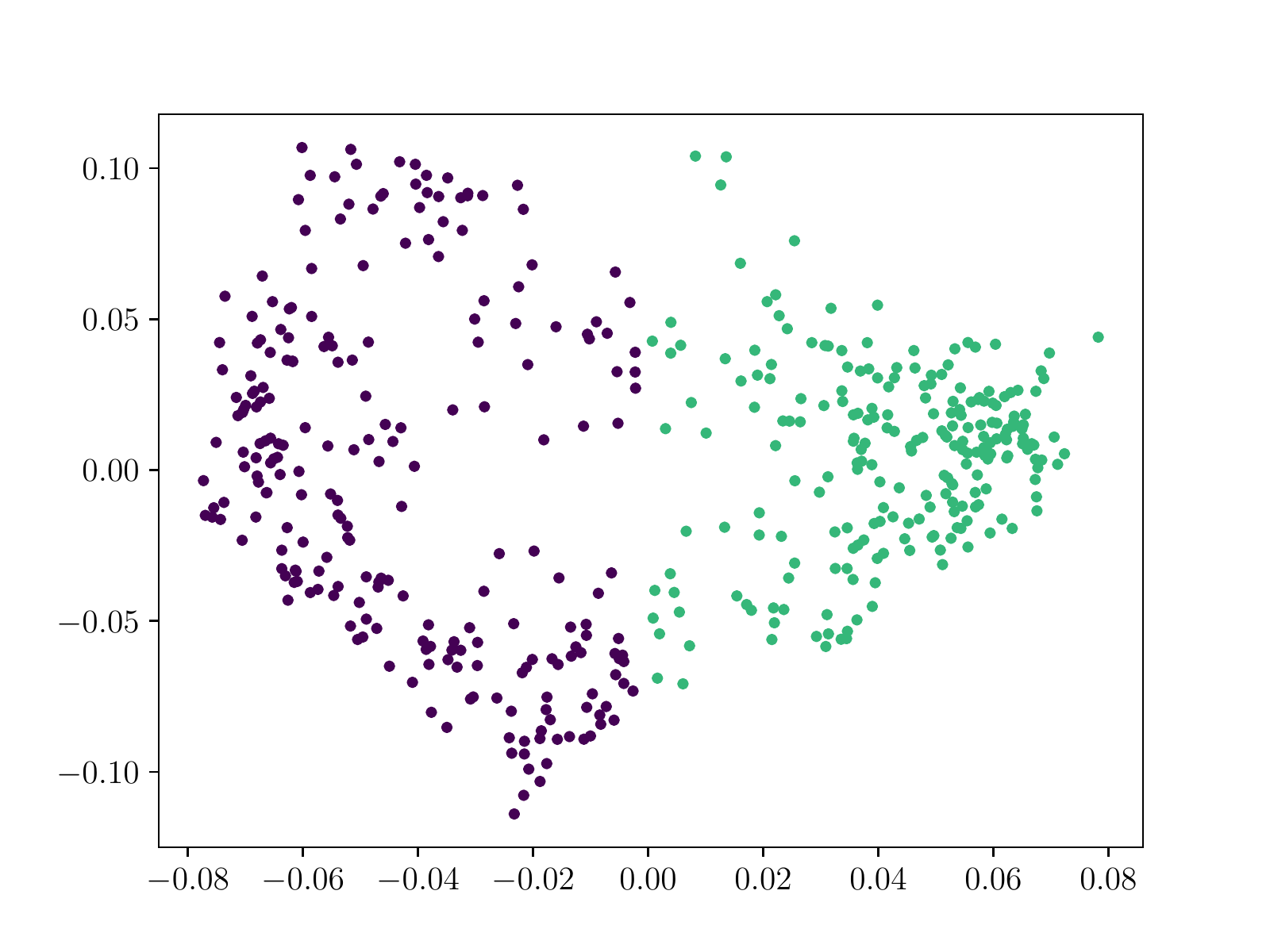} \label{fig:spec_clust}}
\caption[Spectral Clustering]
{Spectral Clustering on Fragment-Level Data\medskip \par \small In (a) we show the spectral embedding of our fragment-level dataset with the points colored based on their true labels of hominin or carnivore. In (b) we show the results of spectral clustering, which runs the $k$-means clustering algorithm on the spectral embedding. The accuracy of the spectral clustering in (b) is $69.8 \%$.}
\end{figure}

In order to visualize our dataset and further explore its structure, we consider here the application of \emph{unsupervised learning algorithms}. Unsupervised learning is a form of machine learning that does not utilize the labels of data points during its training process, and can include algorithms like clustering, dimensionality reduction, and ranking. Here, we used a spectral embedding for dimensionality reduction, and spectral clustering to detect clusters in the dataset. Spectral embeddings offer a way to embed a high dimensional dataset into a low dimensional space that is superior to linear techniques like principal component analysis (PCA). Spectral embeddings build a graph over the dataset based on similarities between datapoints, and the embedding into $k$ dimensions involves computing the first $k$ eigenvectors of the graph Laplacian. Spectral clustering clusters the data by running the $k$-means clustering algorithm on the embedded $k$-dimensional data. We refer to \cite{von2007tutorial} for a tutorial on spectral clustering; we use the specific spectral clustering algorithm proposed in \cite{ng2001spectral}.

In \hyperref[fig:spec_embedding]{Figure} \ref{fig:spec_embedding} we show the spectral embedding of our fragment-level dataset into $k=2$ dimensions, with the points colored by their true labels. We can see a small degree of separation between the classes, though there is significant overlap.  In \hyperref[fig:spec_clust]{Figure} \ref{fig:spec_clust} we show the labels predicted for each point by spectral clustering, which achieved $69.8 \%$ classification accuracy. In particular, the hominin broken fragments were classified at $67.3 \%$ accuracy, while the carnivore broken fragments were classified at $73.4 \%$ accuracy. These unsupervised accuracy values should be viewed as baseline accuracy scores that our fully supervised learning results can be compared to.

\begin{figure}[!t]
\centering
\captionsetup{margin=2cm}
\subfloat[Spectral Embedding]{\includegraphics[width=0.49\textwidth,clip=true,trim= 20 15 30 30]{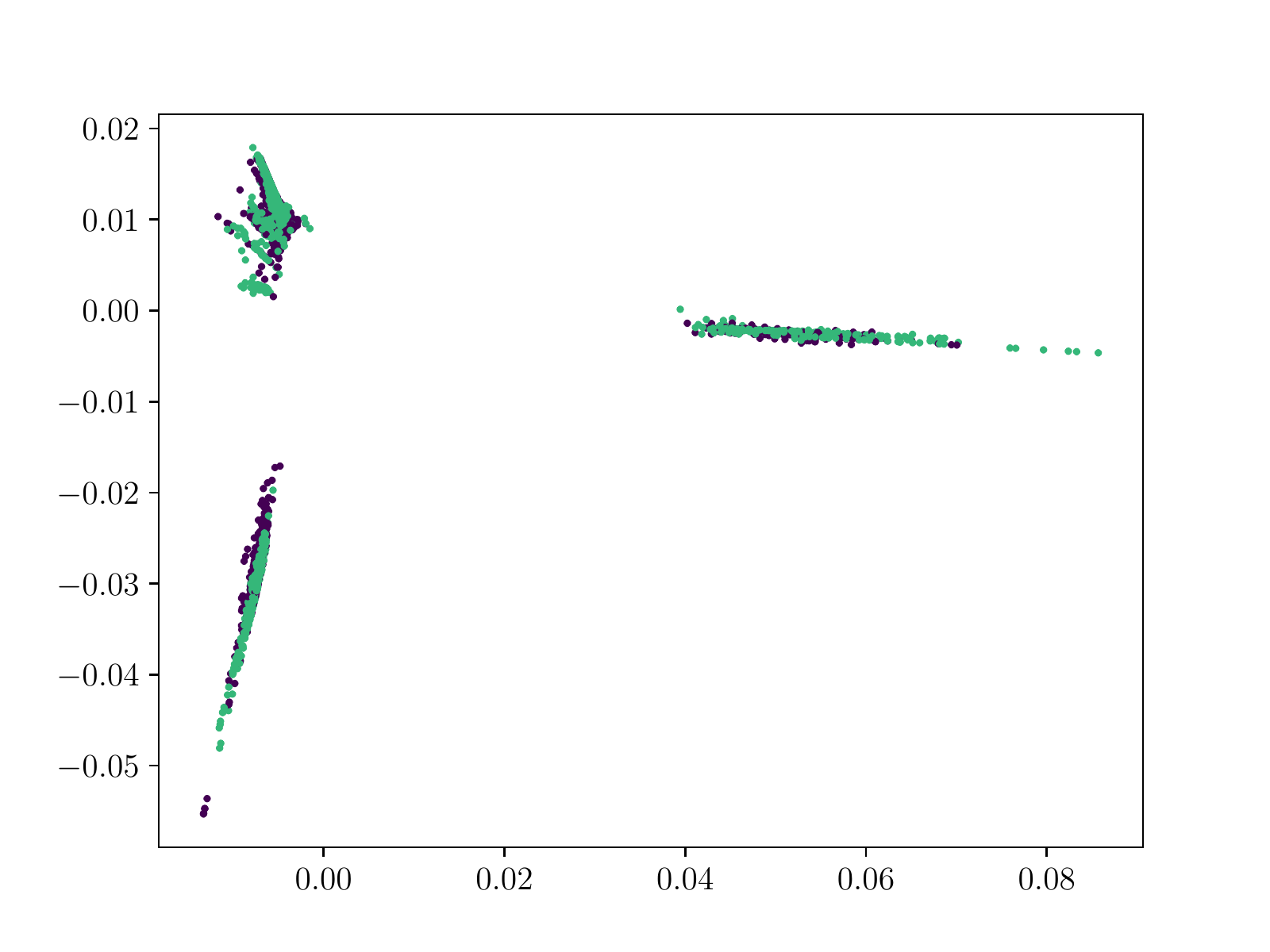} \label{fig:spec_embedding_break}}
\subfloat[Spectral Clustering]{\includegraphics[width=0.49\textwidth,clip=true,trim= 20 15 30 30]{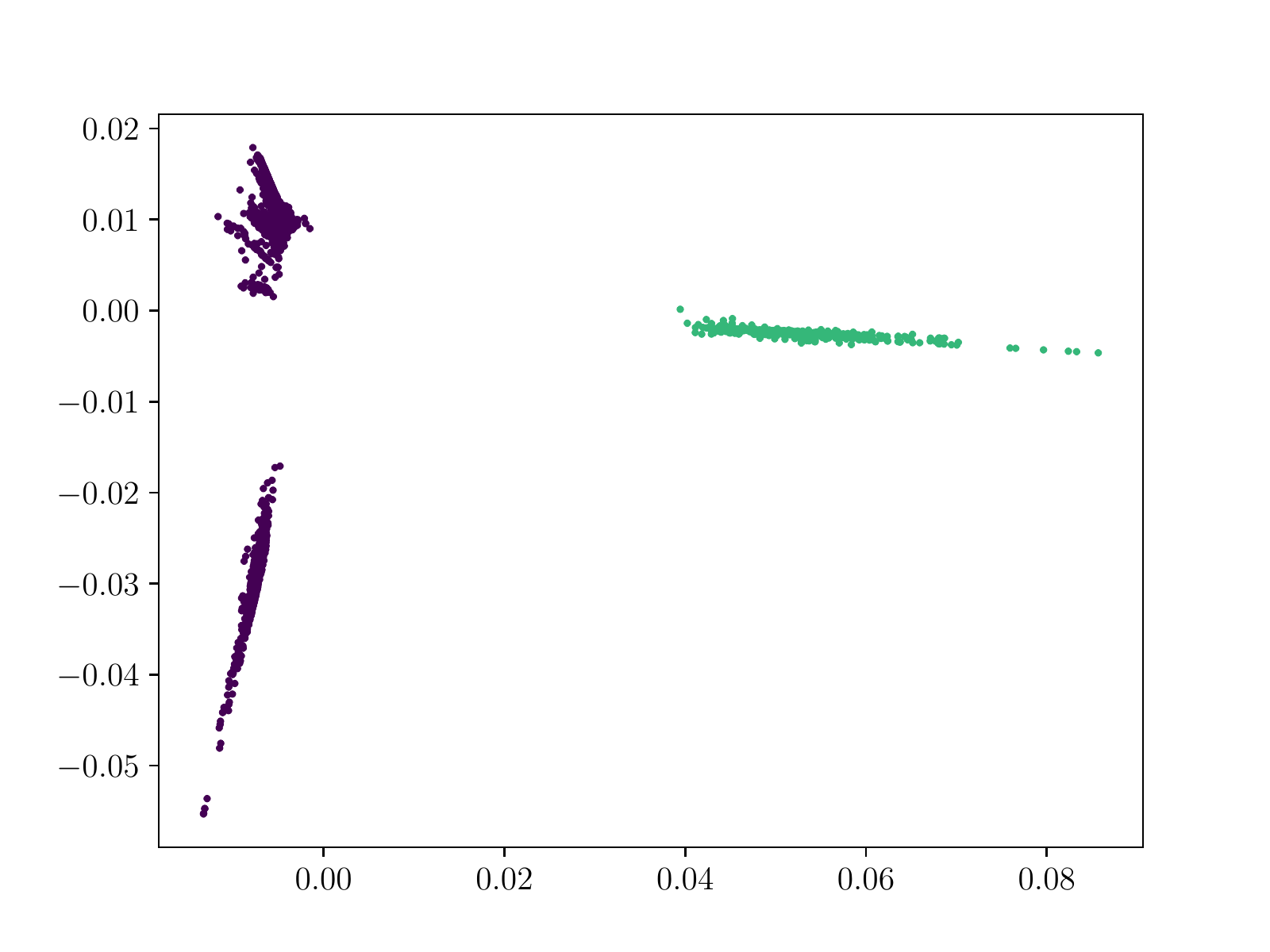} \label{fig:spec_clust_break}}
\caption[Spectral Clustering on Break-Level Data]
{Spectral Clustering on Break-Level Data\medskip \par \small In (a) we show the spectral embedding of our break-level dataset with the points colored based on their true labels of hominin or carnivore. In (b) we show the results of spectral clustering, which achieved a total accuracy of $53.5 \%$.}
\label{fig:}
\end{figure}

In \hyperref[fig:spec_embedding_break]{Figure} \ref{fig:spec_embedding_break} we show the spectral embedding of our break-level dataset into $k=2$ dimensions, with the points colored by their true labels. We see a clear cluster structure here with three well-separated clusters. However, the clusters do not correspond with the classes hominin and carnivore. As a result of this, the spectral clustering on the break-level dataset achieved a total accuracy of $53.5 \%$. Specifically, the hominin breaks were classified at $92.7 \%$ accuracy, while the carnivore breaks were classified at $12.2 \%$ accuracy. These results suggest that the break-level information is useful for classification only when it is compiled through summary statistics at the fragment level, and that considering information on a break-by-break basis yields less useful information for classification.

\section{Comparing our Results to \texorpdfstring{\citealt{moclan2019classifying}}{Moclan et al. 2019}}

In this section, we revisit the machine learning classification results in \citet{moclan2019classifying}, and reanalyze their data using the preceding methods.  We point out significant issues with their applications of machine learning and show that a correct application does not produce the seemingly impressive results they find.  We further compare their data and analysis with ours, as discussed in the preceding section. Finally, we present the results of an experiment with randomized data showing that the issues we identified in \citet{moclan2019classifying} can arbitrarily inflate accuracy scores even when no patterns are present in the data.

\subsection{The \texorpdfstring{\citealt{moclan2019classifying}}{Moclan et al. 2019} Sample}

In our analysis of the results in \citet{moclan2019classifying}, we used their published dataset, which is provided as a \emph{.csv} file in their supplemental information. According to the published $.csv$ file, their sample consists of a total of $1,488$ breaks comprised of $797$ anthropogenic breaks, $177$ breaks created by \textit{Crocuta crocuta} and $514$ breaks created by \textit{Canis lupus}. According to the text in the main article the hyena sample consists of $66$ bones and the wolf sample consists of $237$ fragments. The anthropogenic sample was derived from $40$ bones ($10$ humeri, $10$ radii-ulnae, $10$ femora, and $10$ tibiae) that were broken, resulting in $1,497$ fragments of which they selected $332$. It should be noted that in the first paragraph of their Results Section, they report a total of $881$ hominin produced breaks, $202$ hyena produced breaks, and $610$\footnote{There is a typo. It is written as $61$. However, once the breaks are summed for each fracture plane it is evident that this should be $610$.} breaks produced by wolves. It is not clear why there is a discrepancy between what is written in the main article and what is presented in the supplemental information. Dropping the transverse breaks from analysis does not account for this discrepancy (See \hyperref[tab:moclan]{Table} \ref{tab:moclan}). 

\begin{center}
\begin{threeparttable}[t!]
\vspace{-3mm}
\caption{The \citealt{moclan2019classifying} Sample}
\vspace{-3mm}
\label{tab:moclan}
\vskip 0.15in
\begin{small}
\begin{sc}
\begin{tabular}{lrrrr}
&	Hominin	& Hyena &	Wolf &	Total \\
\toprule
Fragments reported in text	& 332	& 	66	& 	237		& 635 \\
\midrule
Breaks reported in text	& 	881	& 	202		& 610	& 	1693 \\
Breaks reported in SI	& 	797	& 	177		& 514	& 	1488 \\
Difference in reported breaks	& 	84		& 25	& 	96	& 	205 \\
\midrule
longitudinal breaks reported in text	& 	297	 & 91 & 287	& 	675 \\
longitudinal breaks reported in SI	& 	284		& 91	& 	267	& 	642 \\
Difference in reported longitudinal breaks	& 	13	& 	0	& 	20		& 33 \\
\midrule
oblique breaks reported in text	& 	549		& 87	& 	273		& 909 \\
oblique breaks reported in SI	& 	513		& 86	& 	247		& 846 \\
Difference in reported oblique breaks	& 	36	& 	1	& 	26	& 	63 \\
\midrule
transverse breaks reported in text	& 	35	& 	24	& 	50	& 	109 \\
transverse breaks reported in SI	& 	0	& 	0	& 	0	& 	0 \\
Difference in reported transverse breaks	& 	35	& 	24	& 	50	& 	109 \\
\bottomrule \\
\end{tabular} 
\end{sc}
\end{small}
\vskip -0.1in
\end{threeparttable}
\end{center}

Their sample consisted of fragments from animals that weighed $80-200$kg. The fragments from the anthropogenic sample all came from \textit{Cervus elaphus} (red deer). The carnivore samples, which were gathered from a hyena den in Tanzania and a natural park in Spain, included unidentifiable fragments that were not metapodial fragments. They chose not to include metapodia, stating that they were not diagnostic due to the thick cortical bone, citing \citet{capaldo1994quantitative}. The wolf-created sample included fragments from \textit{Cervus elaphus} and \textit{Sus scrofa}. Fragments they chose were $\ge4$ cm in maximum dimension and bore, at minimum, one measurable break. 

\citet{moclan2019classifying}'s dataset contains $12$ variables: Epiphysis, Length, Interval, Number of Planes, Fracture Plane, Plane (Fracture Angle), Type of Angle, $>4$ cm, Notch, Notch A, Notch B, and Notch D (See \hyperref[tab:moclanvars]{Table} \ref{tab:moclanvars}). Epiphysis refers to the presence/absence of some or all of the epiphyseal surface on the fragment. Length refers to the measured length (mm) of the fragment. The Length category is derived by parsing the fragments into bins based on their measured lengths. Number of Planes refers to the number of measurable breaks on the fragment, including transverse breaks. Fracture Plane measures the angle that the fracture plane makes with the longitudinal axis of the bone, as defined by \citet{gifford1989ethnographic}. Though transverse breaks were included in the Number of Planes, only breaks that were longitudinal and oblique were input into the machine learning algorithms. In this count, we interpret ``Plane'' to mean the fracture angle, i.e., the measured angle of transition between the periosteal surface and the break surface along each break curve. In previous studies, the angle was taken at the center along the edge of the break \citep{alcantara2006determinacion, coil2017new, pickering2005contribution}. Here the authors state that it is ``measured at the point of maximum angle. In cases where both acute and obtuse angles are present, the latter was used" \citep[p.3]{moclan2019classifying}. ``Maximum angle'' suggests that the largest angle value was used. However, given the caveat that obtuse angles were used in instances where both acute and obtuse angles were present on a break, the meaning of maximum in this case may refer to the distance from $90^{\circ}$. We assume the measurement was taken with a handheld, contact goniometer; therefore assessing where to take the measurement was likely done, in large part, by eye. (In contrast, our use of the virtual goniometer makes our angle measurements both more accurate and completely reproducible.) Type of angle is derived from parsing the fragments into bins based on their measured angles, where angles $<85^{\circ}$ are categorized as ``acute'', angles between $85^{\circ}-95^{\circ}$ are categorized as ``right'', and angles $>95^{\circ}$ are categorized as ``obtuse''. We interpret ``$>4$ cm'' as meaning the presence or absence of breaks on the fragment that are greater than $4$cm in length. The last four variables identify the presence or absence of notches (in general) on the fragment and the notch types: A, C, and D. \cite[See][for details on notch types]{capaldo1994quantitative}.

\begin{center}
\begin{threeparttable}[t!]
\vspace{-3mm}
\caption{Comparison of Sample Sizes}
\vspace{-3mm}
\label{tab:compare}
\vskip 0.15in
\begin{small}
\begin{sc}
\begin{tabular}{lcccc}
\toprule
& Percussion    & \textit{Crocuta}  & \textit{Canis}    & Total \\
\hline
Moclan  & $332$ ($797$)      &  $66$ ($177$)          & $237$ ($514$)          &  $637$ ($1,488$) \\
Our Experimental Sample & $275$ ($1,651$)          & $188$ ($1,567$)              & $0$($0$)          & $463$ ($3,218$)\\
\bottomrule
\end{tabular}
\begin{tablenotes}[para,flushleft]
\item \footnotesize{The first value is the number of fragments. The value in parentheses is the number of breaks.}\\
\end{tablenotes}
\end{sc}
\end{small}
\vskip -0.1in
\end{threeparttable}
\end{center}

\subsection{Replicating \texorpdfstring{\citealt{moclan2019classifying}}{Moclan et al. 2019}'s Machine Learning Analysis}

\citet{moclan2019classifying} applied six different algorithms (neural networks, support vector machines, $k$-nearest neighbor, random forests, mixture discriminant analysis, and naive Bayes) to their dataset. They ran these tests with and without bootstrapping ($1,000$ times) the raw data \citep[see also][p. 7]{moclan2020identifying}. They separated both the original dataset and the bootstrapped dataset into a $70/30$ training/testing split.  It should be emphasized that bootstrapping prior to splitting the sample into training and test sets is not allowed in machine learning applications because it contaminates the training set with test data; see \hyperref[discussion]{Section} \ref{discussion} for further details. The classification success rate for the original sample ranged between $82\%-89\%$. The classification rates for the bootstrapped sample ranged from $78\%-94\%$. They separated out the breaks according to fracture planes and whether or not the breaks were greater or less than $90^{\circ}$. When applied to the longitudinal fractures with fracture angles $<90^{\circ}$, the classification rates on the original sample were between $75\%-83\%$ and the classification rates for the bootstrapped samples were between $73\%-99\%$. The classification rates for the longitudinal fractures with fracture angles $>90^{\circ}$ showed a success rate of $72\%-82\%$ for the original sample and $81\%-98\%$ for the bootstrapped sample. For the oblique fracture with fracture angles $<90^{\circ}$ classification rates ranged from $68\%-86\%$ for the original sample and $69\%-98\%$ for the bootstrapped sample. Finally, for oblique fractures with fracture angles $>90^{\circ}$, classification rates ranged between $86\%-90\%$ and $89\%-96\%$ (see \hyperref[tab: mocresults]{Table} \ref{tab: mocresults}).

\begin{center}
\begin{threeparttable}[t!]
\vspace{-3mm}
\caption{Summary of \citealt{moclan2019classifying}'s ML Results}
\vspace{-3mm}
\label{tab: mocresults}
\vskip 0.15in
\begin{small}
\begin{sc}
\begin{tabular}{lcc}
\toprule
& Original    & Bootstrapped  \\
\hline
All & $82\%-89\%$ & $78\%-94\%$ \\
Longitudinal $<90^{\circ}$ & $75\%-83\%$ & $73\%-99\%$ \\
Longitudinal $>90^{\circ}$ & $72\%-82\%$ & $81\%-98\%$ \\
Oblique $<90^{\circ}$ & $68\%-86\%$ & $69\%-98\%$ \\
Oblique $>90^{\circ}$ & $86\%-90\%$ & $89\%-96\%$ \\
\bottomrule \\
\end{tabular}
\end{sc}
\end{small}
\vskip -0.1in
\end{threeparttable}
\end{center} 

We replicated their machine learning approach on the entire dataset provided in their supplemental information. Because they did not include specimen information in their dataset, we were unable to replicate our method of splitting by fragment to ensure breaks from the same fragment were not contaminating the testing set. For their dataset, we used repeated $k$-fold cross validation to ensure each data point was included in the test set at least once for each replication. 

We applied random forest, linear support vector machine, neural network, linear discriminant analysis, Gaussian naive Bayes, and $k$-nearest neighbor machine learning algorithms to their dataset. Our use of linear discriminant analysis was a substitution of their mixture discriminant analysis, and we do not expect major differences in algorithm performance.

As in \citet{moclan2019classifying}, we ran the test with and without their inappropriate bootstrapping protocol. However, in the discussion section, we will elaborate on why it is not appropriate to use bootstrapping in this manner when applying machine learning methods, and we chose to run the bootstrapped version here purely for comparison with \citealt{moclan2019classifying}'s work and as a tool for discussion. Additionally, it is important to bear in mind the discrepancies in reported sample sizes mentioned previously in so much that we are making the assumption that they ran the machine learning algorithms on the samples as provided in the supplemental \emph{.csv} file, which could explain any discrepancies with the results we report here.

Unlike \citet{moclan2019classifying}, we did not run tests using subsets of the data based on break plane and fracture angle. This is unnecessary when using machine learning which can parse out which features and relationships among features are useful for classification. Subsetting the data in this way reduces the sample size which exacerbates the issues stemming from the mixing of test data into the training data and the data recording errors. 

As noted above, we have some concerns about \citeauthor{moclan2019classifying}'s data. Some of the variables were corrupted due to what appears to be recording errors. For instance, Epiphysis is a Boolean (present/absent) variable, but $264$ observations were categorized with a $2$, while $78$ observations were categorized as a $3$, plus $37$ observations were categorized as a $4$, while $233$ observations were categorized as ``present'', and $876$ observations were categorized as ``absent''. Likewise, Notch A and Notch C are Boolean variables and in addition to ``present''/``absent'', contained a third value ``2''. Interval length and the type of angle are redundant variables. Indeed, the information contained in these variables is provided by the measured lengths and the measured angles and are therefore unnecessary (see \hyperref[tab:moclanvars]{Table} \ref{tab:moclanvars}).

\begin{center}
\begin{threeparttable}[t!]
\vspace{-3mm}
\caption{Summary of \citealt{moclan2019classifying}'s Variables}
\vspace{-3mm}
\label{tab:moclanvars}
\vskip 0.15in
\begin{footnotesize}
\begin{sc}
\begin{tabular}{lllll}
Variable &	Level &	Type &	Entered Values &	Notes \\
\toprule
Epiphysis &	Frag &	Boolean &	2, 3, 4, Absent, Present &	Corrupted \\
Length (mm) &	Frag &	Numerical &	Whole numbers &	-- \\
&	 &	 &	ranging from 40-267 &	 \\
Interval (length) &	Frag &	Categorical &	Bins: 40-49…90-99,  &	Redundant \\
&	 &	 &	100-149, 150-199, >199 &	 \\
Number of planes &	Frag &	Count &	Whole numbers  &	Transverse breaks  \\
&	 &	 &	ranging from 1-6 &	included in counts \\
Fracture plane &	Break &	Boolean &	Longitudinal, Oblique &	-- \\
Plane/Fracture &	Break &	Numerical &	Whole numbers  &	-- \\
angle &	 &	 &	ranging from $20^{\circ}-161^{\circ}$ &	 \\
Type of angle &	Break &	Categorical &	Acute ($<85^{\circ}$),  &	Redundant \\
 &	 &	 &	Right ($85-95^{\circ}$),  &	 \\
 &	 &	 &	Obtuse ($>95^{\circ}$) &	 \\
$>4$cm &	Frag &	Boolean &	Absent, Present &	-- \\
Notch &	Frag &	Boolean &	Absent, Present &	-- \\
Notch a &	Frag &	Boolean &	2, Absent, Present &	Corrupted \\
Notch c &	Frag &	Boolean &	2, Absent, Present &	Corrupted \\
Notch d &	Frag &	Boolean &	Absent, Present &	-- \\
\bottomrule \\
\end{tabular} 
\end{sc}
\end{footnotesize}
\vskip -0.1in
\end{threeparttable}
\end{center}

Of additional concern are the levels at which the data were collected. Some of the variables were collected at the fragment level. However, the training/testing splits were made at the break level which, as noted above, has the potential for contaminating the training set with data from the test set. It is possible to use data with variables at different levels. However, the data must be split into training/testing splits at the highest level, in this case the fragment level. We will elaborate on this problem in the discussion section. 

Given these data challenges, we ran additional tests on their data, but, first we dropped all corrupted, redundant, and fragment-level variables. We were unable to use the fragment level variables here because the \emph{.csv} file did not identify from which fragment each break was derived so we could only split the sample at the break level. Cleaning the data reduced the variables to break plane and fracture angle. In the first iteration, we maintained the three groups in order to compare the results of the properly run machine learning test against the results of the previous tests. We then pooled the carnivore samples in order to compare the results with our experimental sample.   

Despite the inherent issues with the dataset, we were able to run the machine learning algorithms on the entire dataset with and without bootstrapping to offer a point of comparison between the results that \citet{moclan2019classifying} achieved and the results one can expect to achieve when the data are cleaned and the machine learning algorithms are properly applied. When bootstrapping was used we achieved successful classifications rates ranging from $86.93\%-95.52\%$ with standard deviations ranging from $1.33\%-2.35\%$ (see \hyperref[tab:Moclanbootstrap]{Table} \ref{tab:Moclanbootstrap}). These results are similar to that achieved by \citet{moclan2019classifying}.

It is expected that the Random Forest algorithm will perform  well on bootstrapped data, as it is able to leverage the duplication contamination of the testing dataset due to the nature of decision tree classifiers. The $k$-nearest neighbor algorithm also performs well when we set $k = 1$, which outperforms all other $k$ for this level of bootstrapping. This is precisely because the bootstrapping replicates datapoints and ensures that every datapoint appears multiple times in both the training and testing set, so the nearest neighbor is always the duplicated point with the correct label. Without bootstrapping, classification rates dropped, ranging from $80.74\%-87.83\%$ with standard deviations ranging from $2.60\%-3.15\%$; see \hyperref[tab:Moclannoboots]{Table} \ref{tab:Moclannoboots}. Again, these results are similar to those achieved by \citet{moclan2019classifying}.

\begin{center}
\begin{threeparttable}[t!]
\vspace{-3mm}
\caption{ML Results Using Moclan's Data \\ (With Bootstrapping)}
\vspace{-3mm}
\label{tab:Moclanbootstrap}
\vskip 0.15in
\begin{small}
\begin{sc}
\begin{tabular}{lcc}
\toprule
 & Mean  & Standard  \\
Algorithm &  Accuracy &  Deviation \\
\hline
Random forest (RF) & $95.52\%$ & $1.33\%$ \\
Support vector machine (SVM) – linear & $87.80\%$ & $2.12\%$ \\
Neural network (NN) & $87.56\%$ & $2.08\%$ \\
Linear discriminant analysis (LDA) & $86.93\%$ & $2.10\%$ \\
Gaussian naive Bayes (GNB) & $81.24\%$ & $2.35\%$ \\
$k$-nearest neighbor ($k$-NN) & $94.25\%$ & $1.47\%$ \\
\bottomrule
\end{tabular}
\begin{tablenotes}[para,flushleft]
\item \footnotesize{300 repetitions, 10 folds, 1,000 boostraps}\\
\end{tablenotes}
\end{sc}
\end{small}
\vskip -0.1in
\end{threeparttable}
\end{center}

\begin{center}
\begin{threeparttable}[t!]
\vspace{-3mm}
\caption{ML Results Using Moclan's Data \\ (Without Bootstrapping)}
\vspace{-3mm}
\label{tab:Moclannoboots}
\vskip 0.15in
\begin{small}
\begin{sc}
\begin{tabular}{lcc}
\toprule
 & Mean  & Standard  \\
Algorithm &  Accuracy &  Deviation \\
\hline
Random forest (RF) & $87.83\%$ & $2.60\%$ \\
Support vector machine (SVM) – linear & $86.60\%$ & $2.60\%$ \\
Neural network (NN) & $82.57\%$ & $3.03\%$ \\
Linear discriminant analysis (LDA) & $86.16\%$ & $2.71\%$ \\
Gaussian naive Bayes (GNB) & $80.74\%$ & $3.15\%$ \\
$k$-nearest neighbor ($k$-NN) & $82.32\%$ & $3.04\%$ \\
\bottomrule
\end{tabular}
\begin{tablenotes}[para,flushleft]
\item \footnotesize{300 repetitions, 10 folds}\\
\end{tablenotes}
\end{sc}
\end{small}
\vskip -0.1in
\end{threeparttable}
\end{center}

Due to the fragment level contamination of the data, Random Forest is still expected to do well. We used $k=25$ for the $k$-nearest neighbor algorithm, which did decently, but could potentially be optimized by modifying the value of $k$. We used a neural network with three hidden layers of size $100$, $200$, and $400$, respectively, and trained it in a similar way as we described earlier in the paper.

These results are extremely appealing, however, given the errors in data collection and the application of bootstrapping, the results are unreliable. Therefore, we cleaned the data by dropping all the variables that had recording errors and dropping all the fragment level variables. We were not able to incorporate fragment level variables because the published dataset does not contain fragment labels, and thus we were unable to divide the set into training and test sets at the fragment level and we were unable to classify at the fragment level. 

We ran the machine learning algorithms initially with three labels: hominin, hyena, and wolf. Then we repeated the process after pooling the hyena and wolf into a carnivore class so that we could make direct comparisons with the results we achieved using our experimental data. 


Using three labels, one can expect a classification accuracy of approximately $33\%$ by random guessing. We achieved classification rates between $53.74\%-57.59\%$ with standard deviations ranging from $3.73\%-4.29\%$ (see \hyperref[tab: moclan3]{Table} \ref{tab: moclan3}). Though slightly better than what can be expected with random choice, the proper application of machine learning results, not surprisingly, in a dramatic decline in the overall accuracy. 

\begin{center}
\begin{threeparttable}[t!]
\vspace{-3mm}
\caption{ML Results Using Moclan's Data \\ (Break Level Only - 3 Actors)}
\vspace{-3mm}
\label{tab: moclan3}
\vskip 0.15in
\begin{small}
\begin{sc}
\begin{tabular}{lcc}
\toprule
 & Mean  & Standard  \\
Algorithm &  Accuracy &  Deviation \\
\hline
Random forest (RF) & $55.00\%$ & $3.83\%$ \\
Support vector machine (SVM) – linear & $53.74\%$ & $4.29\%$ \\
Neural network (NN) & $58.33\%$ & $3.92\%$ \\
Linear discriminant analysis (LDA) & $57.59\%$ & $3.77\%$ \\
Gaussian naive Bayes (GNB) & $54.92\%$ & $4.05\%$ \\
$k$-nearest neighbor ($k$-NN) & $56.08\%$ & $3.73\%$ \\
\bottomrule
\end{tabular}
\begin{tablenotes}[para,flushleft]
\item \footnotesize{300 repetitions, 10 folds}\\
\end{tablenotes}
\end{sc}
\end{small}
\vskip -0.1in
\end{threeparttable}
\end{center}

Without the fragment level contamination to leverage, the Random Forest algorithm no longer leads in mean accuracy, and other algorithms that aren't based on Decision Tree methods overtake it.

When carnivores were pooled the classification mean accuracy ranged from $59.25\%-64.21\%$ with standard deviations ranging from $3.70\%-4.41\%$ (see \hyperref[tab: moclan2]{Table} \ref{tab: moclan2}). Our mean accuracy, using only break-level data, was, on average, a bit higher ($64\%$ as opposed to $61\%$), had a slightly wider range ($57.18-68.64\%$) and slightly higher standard deviations ($4.22\%-5.06\%$). Using both break level and fragment level variables to classify fragments, our classification rates improved with mean accuracy ranging from $72.82\%-79.27\%$ and lower standard deviations $3.42\%-3.94\%$

\begin{center}
\begin{threeparttable}[t!]
\vspace{-3mm}
\caption{ML Results Using Moclan's Data \\ (Break Level Only - 2 Actors)}
\vspace{-3mm}
\label{tab: moclan2}
\vskip 0.15in
\begin{small}
\begin{sc}
\begin{tabular}{lcc}
\toprule
 & Mean  & Standard  \\
Algorithm &  Accuracy &  Deviation \\
\hline
Random forest (RF) & $61.19\%$ & $3.70\%$ \\
Support vector machine (SVM) – linear & $61.20\%$ & $4.41\%$ \\
Neural network (NN) & $64.21\%$ & $3.76\%$ \\
Linear discriminant analysis (LDA) & $62.80\%$ & $3.78\%$ \\
Gaussian naive Bayes (GNB) & $59.25\%$ & $3.73\%$ \\
$k$-nearest neighbor ($k$-NN) & $61.63\%$ & $3.73\%$ \\
\bottomrule
\end{tabular}
\begin{tablenotes}[para,flushleft]
\item \footnotesize{300 repetitions, 10 folds}\\
\end{tablenotes}
\end{sc}
\end{small}
\vskip -0.1in
\end{threeparttable}
\end{center}

\subsection{An experiment with randomized data}
\label{sec:random}

In order to further illuminate the issues we have identified from \cite{moclan2019classifying} with bootstrapping and break-level train-test splits, we applied machine learning algorithms to a random dataset that we constructed of a similar size to the dataset used in \cite{moclan2019classifying}. Our random synthetic dataset has $200$ fragments, each with $7$ breaks, yielding $1400$ breaks, which is comparable to the $1488$ used in \cite{moclan2019classifying}. Each fragment is assigned $34$ random numerical features, and each break is assigned $6$ random numerical features. The number of features is similar to the number of break and fragment-level features used in \cite{moclan2019classifying}, after the categorical features are converted to numerical features through one-hot encodings. This yields a dataset of $1400$ breaks, each of which has $40$ numerical features ($34$ fragment-level and $6$ break-level). Each fragment is then assigned a label of $0$ or $1$ uniformly at random, and that label is transferred to the break. We emphasize that this dataset is constructed completely at random, so there is no information in the dataset from which a machine learning algorithm can learn. Any proper application of machine learning should achieve on average $50\%$ classification accuracy. 

\begin{center}
\begin{threeparttable}[t!]
\vspace{-3mm}
\caption{Results of the randomized machine learning experiment.}
\vspace{-3mm}
\label{tab:rand}
\vskip 0.15in
\begin{small}
\begin{sc}
\begin{tabular}{lccc}
\toprule
& Break-Level & Frag-Level Split & Frag-Level\\ 
Algorithm & Split &  with Bootstrapping & Split \\ 
\midrule
LDA& 65.1 (4.1)& 96.7 (4.1)& 50.1 (6.9)\\ 
Random Forest& 100.0 (0.0)& 100.0 (0.0)& 50.7 (7.3)\\ 
Linear SVM& 66.5 (3.7)& 100.0 (0.0)& 50.2 (7.4)\\ 
RBF SVM& 99.6 (0.7)& 100.0 (0.0)& 49.6 (7.0)\\ 
Nearest neighbor& 100.0 (0.1)& 100.0 (0.0)& 50.0 (6.4)\\ 
Neural Network& 63.3 (3.8)& 100.0 (0.0)& 50.7 (5.8)\\ 
\bottomrule
\end{tabular}
\begin{tablenotes}[para,flushleft]
\item \footnotesize{100 repetitions, standard deviation in parentheses}\\
\end{tablenotes}
\end{sc}
\end{small}
\vskip -0.1in
\end{threeparttable}
\end{center}

We applied six machine learning algorithms to this dataset: linear discriminant analysis (LDA), random forest, linear SVM, SVM with radial basis function (RBF) kernel, a nearest neighbor classifier, and a neural network. We considered three separate experiments. First, we performed machine learning with a train-test split at the break level, as was done in \cite{moclan2019classifying}. Second, we bootstrapped the fragment-level data $100$ times before doing a fragment-level train-test split. Third, we applied machine learning correctly by simply performing the train-test split at the fragment level. We show the results of these three experiments in \hyperref[tab:rand]{Table} \ref{tab:rand}. All experiments were averaged over $100$ trials of randomized train-test splits,  and we report the mean accuracy with the standard deviation in parentheses. 

We can see that both the break-level train-test split, and bootstrapping $100$ times, which is less than the $1000$ times used in \cite{moclan2019classifying}, both yield accuracy scores near $100\%$ for many algorithms. These accuracy scores are artificially inflated, since the dataset is random. Any properly applied machine learning algorithm can achieve no better than $50\%$ accuracy on average. We note that some algorithms, like LDA, linear SVM, and neural networks, do not achieve very high accuracies with the break-level split. The duplication of datapoints in the training and testing set can only be exploited by machine learning models that can easily overfit. Linear SVM and LDA are very low complexity models that cannot easily overfit, without a larger amount of duplication, like in the second experiment which was bootstrapped $100$ times. The break-level split can be thought of as bootstrapping $7$ times, since there are $7$ breaks per fragment. Neural networks have the capacity to overfit, due to the number of parameters in the model, but the specific techniques used in training, such as stochastic gradient descent, offer some protection against overfitting, though the acutal mechanisms by which this occurs are currently an open problem in deep learning \citep{zhang2021understanding}.

\section{Discussion}
\label{discussion}

When applied properly, machine learning can be an effective tool that can be used to advance our understanding of human evolution through the analysis of fracture patterns. However, the data need to be clean (i.e. properly and consistently recorded),  there needs to be a clear understanding of the ways in which the data can be properly used, as well as a sense for the quality of the data.  The quality of the data can be assessed from the perspective of how much information can be conveyed as well as replicability and transparency. Independent replication is key for testing the ability of the data to answer the question of interest. 
The quality of the features presented here surpasses what has been achieved in prior work in terms of the amount of information that is extracted, transparency, and  replicability. 

Two limitations to this study are the sample size and the sample distribution in respect to species and skeletal element which are not uniformly balanced. These are fundamental limitations that we must continually address within the discipline and is continually improved through the expansion of the samples as research continues. More importantly, we have set a reliable groundwork from which to build future research by properly applying machine learning methods on more robust feature sets. 

One of the primary challenges is how to subdivide the broken surface of a fragment into separate breaks. This issue has been acknowledged in the published research since the 1970s \citep{biddick1975quantifying, davis1985taphonomic, pickering2005contribution, bunn1982meat, bunn1989diagnosing}. The selection of break endpoints and the measurement of angles using the virtual goniometer results in visualizations that clearly communicate how we chose to subdivide the fragments in our sample, which can then be the basis for conversations that can move beyond the recognition of the challenge toward the development of a quantifiable, replicable convention. These visualizations also offer a level of transparency by which other researchers can independently evaluate the extent to which an established protocol was followed. 

Additionally, the points along the edge were used to create two measures of distance for each break and, given that the $x$, $y$, and $z$ coordinates for each point are recorded, the measurements can be replicated with high precision. This is a substantial improvement over using calipers which are subject to variations resulting from the physical interaction of the measuring implement and the fragment which leads to higher error ranges. And, the arc length cannot be measured with the calipers alone. Likewise, using the bounding box dimensions extracted from 3D models using a Python script for measures of fragment length, width, and depth are more precise than those obtained using calipers, while additional features such as volume and surface area can be extracted as well. This also holds true for the angle measurements; the precision, accuracy, and reproducibility of the virtual goniometer far surpasses that of the handheld contact goniometer \citep{yezzi2021virtual}. And, multiple measurements were taken along the entire break curve which provides substantially more information about the fracture angles. 

Another commonly used variable is the break plane which has been defined as the angle of the break relative to the longitudinal axis of the bone. This is a qualitative, categorical variable with three possible labels: longitudinal, transverse, and oblique. Here we have replaced that with a continuous variable that can be calculated with precision using the points along the break curve and the principal axis of the bone fragment. 

Though we have not completely abandoned the use of qualitative variables, we have used a substantial number of variables that offer precise quantitative values and can be easily replicated. Transparency is promoted through the recording of metadata and visualizations of the data on the 3D models. Because most of the data are automatically extracted using programming scripts or as a built-in feature to plug-ins using a graphical user interface, the potential for errors during data recording are minimized.  As a consequence, we are satisfied that our dataset is clean, with no recording errors, is transparent and replicable.  Moreover, we have improved the quality of the features relative to previous work by extracting features that offer more information about the bone fragment than ever before.

The next consideration then, is how to properly use the data when employing machine learning. Firstly, bootstrapping is used in machine learning (e.g. random forest). However, when bootstrapping is used, it is a hyperparameter built into the algorithm and not something that should be done separately. It is an egregious error to bootstrap a sample prior to subdividing it into training and test sets. This is because sampling with replacement followed by subdivision of the bootstrapped sample will lead to duplicates shared between the training and testing sets, which, of course, leads to exact matches and can then falsely inflate the accuracy of the model. In a proper application of machine learning, the testing set is completely independent of the training set, and is not used in any way during training. This allows the test accuracy to serve as an unbiased measure of how well the model generalizes to new data.

Another important consideration is the scale at which data are classified relative to the scale of the features used in classification and how the data are split. In this case, there are two scales: the entire fragment and the individual breaks on each fragment. Once again, this is an issue of contaminating the training set with duplicates from the test set. To offer a simplified hypothetical example, consider a dataset in which there is only one fragment that is exactly $10$ cm long that has $5$ break faces, and the goal is to classify all the breaks in the dataset. If $3$ of those breaks are added to the training set and the other $2$ land in the test set, then the model will identify $10$ cm as belonging to a specific label. The model will be built from this and when the test samples are used to validate the model, they will most likely classify on the basis of this feature alone. The key here is to ensure that the data are split at the appropriate level (i.e. fragment level) so as not to contaminate the training set with information from the testing set. Algorithms with many degrees of freedom (like Decision Trees or Random Forest) can easily discern a correspondence between a variable like dimension and actor, provided each fragment has a different value for dimension. The algorithm memorizes that relationship, which is not useful when new fragments are introduced to the model. \citet{moclan2019classifying} should have classified at the fragment level, used only variables that were specific to the breaks, or split the data at the fragment level. We chose to use two levels of data in our classification. However, we also chose to split the data at the fragment level and classify entire fragments as opposed to individual breaks. Though our results were not as compelling as those produced by \citet{moclan2019classifying}, they are based on proper applications of machine learning and therefore are valid and reliable results. Our results are promising in that they are able to discriminate between the agents of breakage. 

\section{Conclusion and Future Research}

The driving motivation for this line of research is to better understand early hominin evolution and how marrow exploitation factored into subsistence patterns that influenced that evolution. The results achieved here through the proper application of machine learning show promise for identifying agents of bone breakage in the archaeological record, which in turn will be useful for addressing higher level questions focused on subsistence. Future research directions that can further develop and strengthen this approach include, first and foremost, increasing the comparative samples and testing these methods on models that emulate the effects of taphonomic processes, e.g., erosion. The hominin-carnivore debate has, as the name suggests, been treated as a binary problem when, in fact, other agents can break bones, e.g., geological processes, so including samples derived from other methods of bone breakage will be an important next step. 

It is also important to understand how machine learning algorithms make their predictions, to shed some light on which features are important for classification, and which are redundant or can be viewed as noise. While this interpretability problem in machine learning is well-documented and is often cited as an issue in the field, key factors in the decision making can be determined by probing the machine learning models in appropriate ways, and doing so in the anthropological context may yield interesting and important information. It would also be interesting to understand better the clustering structure of our dataset, and what characterizes the clusters in, for example, \hyperref[fig:spec_embedding_break]{Figure} \ref{fig:spec_embedding_break}. 

Another interesting avenue of research would be to explore fragments that consistently classify poorly to examine in more detail the sources of equifinality. Success in this direction would aid researchers in either finding features sets that can address equifinality or identify specimens that cannot be used for classification. This holds promise for much higher resolution analyses of paleoanthropological sites whereby zooarchaeological collections are not restricted to analysis at the assemblage level. The classification of individual fragments opens possibilities for spatial analysis and reconstructing more complex details of the dynamic systems and interactions that formed important paleoanthropological sites. 

The present work represents a strong first step in the direction of leveraging richer datasets and the application of advanced computational tools that hold promise for the future of taphonomic research in paleoanthropology.


\bibliographystyle{apacite}
\bibliography{yezziwoodley.bib}

\end{document}